%% file: main.tex
\documentclass{article}
\usepackage{ijcai23}

\usepackage{times}
\usepackage{soul}
\usepackage{url}
\usepackage[hidelinks]{hyperref}
\usepackage[utf8]{inputenc}
\usepackage[small]{caption}
\usepackage{graphicx}
\usepackage{color}
\usepackage{amsmath}
\usepackage{amsthm}
\usepackage{booktabs}
\usepackage{algorithm}
\usepackage[switch]{lineno}
\usepackage{amssymb}
\usepackage{algpseudocode}
\usepackage{multirow}
\usepackage{float}
\usepackage{subcaption}
\usepackage[title]{appendix}
\usepackage{xcolor}

\newtheorem{theorem}{Theorem}

\title{Fulfilling Formal Specifications ASAP by Model-free Reinforcement Learning}

\author{
Mengyu Liu$^{1*}$\and
Pengyuan Lu$^{2*}$\and
Xin Chen$^3$\and
Fanxin Kong$^1$\and
Oleg Sokolsky$^2$\And
Insup Lee$^2$
\affiliations
$^1$Syracuse University\\
$^2$University of Pennsylvania\\
$^3$University of Dayton
\emails
mliu71@syr.edu,
pelu@seas.upenn.edu,
xchen4@udayton.edu,
fkong03@syr.edu,
\{sokolsky,lee\}@seas.upenn.edu
}

\newtheorem{definition}[theorem]{Definition}

\newcommand\E{\mathbb{E}}
\newcommand\statespace{\mathcal{S}}
\newcommand\actionspace{\mathcal{A}}
\newcommand\signal{\bar{s}}
\newcommand\state{s}
\newcommand\action{a}
\newcommand\property{\varphi}
\newcommand\policy{\pi} 
\newcommand\policyparam{\theta}
\newcommand\sysmodel{p}
\newcommand\reward{r}
\newcommand\criticval{v} 
\newcommand\criticvalparam{w}
\newcommand\qs{\rho} 
\newcommand\ac{AC} 

\usepackage[normalem]{ulem}

\begin{document}

\maketitle
\def\thefootnote{*}\footnotetext{These authors contributed equally to this work}

\begin{abstract}
\input{sections/0_abstract.tex}
\end{abstract}

\input{sections/1_introduction.tex}

\input{sections/2_preliminaries.tex}
\input{sections/3_problem.tex}
\input{sections/4_framework.tex}

\input{sections/5_experiments.tex}

\input{sections/6_discussion.tex}
\input{sections/7_conclusion.tex}

\bibliographystyle{named}

\input{sections/references}
\newpage
\input{sections/8_appendix.tex}

\end{document}

%% file: sections/0_abstract.tex
We propose a model-free reinforcement learning solution, namely the ASAP-Phi framework, to encourage an agent to fulfill a formal specification ASAP. The framework leverages a piece-wise reward function that assigns quantitative semantic reward to traces not satisfying the specification, and a high constant reward to the remaining. Then, it trains an agent with an actor-critic-based algorithm, such as soft actor-critic (SAC), or deep deterministic policy gradient (DDPG). Moreover, we prove that ASAP-Phi produces policies that prioritize fulfilling a specification ASAP. Extensive experiments are run, including ablation studies, on state-of-the-art benchmarks. Results show that our framework succeeds in finding sufficiently fast trajectories for up to 97\% test cases and defeats baselines.

%% file: sections/1_introduction.tex
\section{Introduction}
\label{sec:introduction}

One significant use of model-free reinforcement learning is to train an agent to fulfill a formal specification without concrete knowledge of the environment. However, sometimes we need agents to accomplish some tasks that the state-of-the-art specification languages cannot express. For example, the agent needs to reach a goal as soon as possible (ASAP), motivated by the scenario where a real-time cyber-physical system is recovering from attacks. Unfortunately, although ``accomplishing a task ASAP'' is often a standard request in real-time control, how to develop a control policy based on data-driven or black box dynamics is still not well studied.

Formal languages are often used to describe control goals. For example, the reach-avoid problem on discrete and hybrid systems can be defined in linear temporal logic (LTL) \cite{vardi1996automata} and signal temporal logic (STL) \cite{raman2014model,deshmukh2017robust} respectively. Most of the existing controller synthesis approaches requires to have a formal model~\cite{sadraddini2018formal,donze2015blustl,chang2019neural}, a simulator~\cite{ljung1998system,degris2012model,ma2021data}, or an identification~\cite{qin2022sablas} of the system dynamics. Although the use of reinforcement learning (RL) in controller synthesis for black box systems has already been intensively studied~\cite{ding2011mdp,fu2014probably,li2017reinforcement,li2018policy,hamilton2022training}, very few work addresses the ASAP property.


Our goal is to develop real-time controllers that fulfills a formal specification ASAP for black box systems. This solution will provide a way to guide a controller synthesis procedure when formal languages are incapable of expressing certain information about the task, and in our case, the request for ASAP. In addition, once found, the solution can be applied to various scenarios. For example, it can enhance security of cyber-physical systems by solving the real-time recovery problem when the environment dynamics is unknown. The recovery problem asks an agent to find a control sequence to steer the system to a pre-defined target state within the shortest time. Existing methods leverage formal models of the system dynamics, such as ordinary differential equations (ODE), to synthesize a controller ~\cite{kong2018cyber,zhang2020real,zhang2021real}, but to our knowledge, no work has addressed black box dynamics so far.

Our framework is called \emph{ASAP-Phi}. It trains an agent to satisfy a given STL property with the ASAP request. Furthermore, the framework consists of: (1) constructing a particular reward function that encourages satisfying the specification ASAP, and (2) using the reward function to train a control policy on an actor-critic-based model-free RL algorithm. We prove that our ASAP-Phi framework encourages ASAP property satisfaction because any optimal policy produced must be ASAP, and design an algorithm to apply this framework to the recovery problem, to reach target sets ASAP while avoiding obstacles. The effectiveness of our approach is evaluated based on various state-of-the-art benchmarks such as OpenAI Gym \cite{stable-baselines3}, and the ablation studies to compare ASAP-Phi with alternative methods for deriving ASAP property. Overall, our contributions include
\begin{enumerate}
    \item A model-free RL framework named ASAP-Phi, that takes in the information of state space, action space and a specification we want the agent to satisfy ASAP, and outputs a control policy to achieve so, without any knowledge of the environment dynamics.
    \item A proof that shows any optimal policy produced by ASAP-Phi indeed achieves a property ASAP.
    \item Experiments demonstrate that ASAP-Phi has up to $97\%$ high rate to succeed in tasks that requires ASAP behaviors and beats baseline methods, as well as ablation studies that show its advantages over alternative approaches.
\end{enumerate}



%% file: sections/2_preliminaries.tex
\section{Preliminaries}
\label{sec:preliminaries}



\subsection{Notations}
\label{subsec:notations}

In the paper, we use $\statespace$ to denote the \emph{state space} of a system. A \emph{trajectory} (or trace) is a finite sequence of states $\bar{s} = s_1,\dots,s_k$ such that $s_i\in \statespace$ for all $i = 1,\dots,k$. We only consider discrete trajectories of the system and we sometimes use $s_t$ to denote the system state at the time $t$.
The notation $\actionspace$ stands for the set of \emph{actions} that can be applied to the system. The rest of the notations will be introduced along the paper upon their appearances.

\subsection{Actor-critic RL}
\label{subsec:actor_critic_rl}

Actor-critic RL is a class of model-free RL algorithms that maintains and updates two models, which represents the learned policy (actor) $\policy$ and the estimated value function (critic) $\criticval$, respectively. The training is to search for an optimal $\criticval^*$ that estimates the expected value, and an optimal $\policy^*$ that maximizes the estimated value. Formally,
\begin{equation}
    \label{eq:ac_bellman}
    \begin{split}
        &\criticval_\policy^*(\state_t) = \E_\policy[\reward(\state_t) + \gamma\criticval^*(\state_{t+1})|\state_t] \\
        &\policy^*(\state_t) = \arg\max_{\policy}\criticval_\policy^*(\state_t).\\
    \end{split} 
\end{equation}
where $r$ is the reward function, $s_t, s_{t+1}$ are the current and next system state respectively. $\E_\policy$ denotes the expectation under the policy $\policy$, and $0 < \gamma < 1$ is the discount factor.
Finding the solutions to Equation \eqref{eq:ac_bellman} is the goal of actor-critic RL \cite{peters2005natural}. Here, the optimal $\criticval^*$ is a solution to the recursive Bellman equation that depends on the policy selected, and the optimal $\policy^*$ aims to maximize the expectation.

There are various methods to implement an actor-critic RL. For example, soft actor-critic (SAC) leverages entropy of distributions to regularize the value estimation \cite{haarnoja2018soft}. Deep deterministic policy gradient (DDPG) treats policies as deterministic and uses delayed copies of actor and critic models to improve robustness against training data variance \cite{silver2014deterministic}. We can find well developed methods from the Stable Baselines module \cite{stable-baselines3}.

\subsection{Formal Specifications and STL}
\label{subsec:formal_specs}

In the area of formal methods, formal specifications are unambiguous descriptions of the tasks that we want agents to accomplish. They can be expressed in multiple formal languages, for example, linear temporal logic (LTL \cite{vardi1996automata} and signal temporal logic (STL) \cite{raman2014model,deshmukh2017robust}. In this paper, we focus on STL. An STL formula can be defined using the following syntax:
\begin{equation}
    \label{eq:stl_grammar}
    \property ::= \top \mid f(\signal) < 0 \mid \neg\property \mid \property_1 \land \property_2 \mid \property_1 U_{[t_1, t_2]} \property_2
\end{equation}
such that $\top$ is tautology, and $f$ is a function that maps a trajectory (or equivalently, a signal, in the context of STL) 
$\signal$ to a real number. $U$ is the until operator, e.g., $\property_1 U_{[t_1, t_2]} \property_2$ indicates that $\property_2$ must hold at some time $t\in [t_1,t_2]$ and $\property_1$ must always hold before that. The time interval $[t_1,t_2]$ is interpreted as $\{t_1,t_1+1,\dots,t_2\}$.
Based on this grammatical structure, STL has guided control synthesis in the field of optimal control \cite{sadraddini2018formal,donze2015blustl,raman2014model}. However, we must admit there is a certain lack of expressiveness of STL and other formal languages, such as they cannot express some common requirements to humans like "as soon as possible".

\subsection{Specification-guided RL}
\label{subsec:spec_guided_rl}
Existing work in RL tried multiple ways to leverage formal specification to guide agent training. Specifically, there are two major approaches in the current state-of-the-art. First, some construct a product Markov decision process (MDP) to combine the underlying MDP and the specification, which is transformed into a finite state machine \cite{fu2014probably,ding2011mdp}. Alternatively, less guaranteed but more flexible approach is to construct a reward function based on the formal specification. For example, quantitative semantics is a real-valued measure $\qs$ on how much a signal $\signal$ satisfies an STL property $\property$ at time $t$ \cite{hamilton2022training}. 
We restate 
its definition in our notations as follows.
\begin{equation}
    \label{eq:qs_definition}
    \begin{split}
    &\qs(\signal, t, (f(\signal) < d)) = d - f(\state_t)\\
    &\qs(\signal, t, \neg\property) = -\qs(\signal, \property, t)\\
    &\qs(\signal, t, \property_1 \land \property_2) = \min(\qs(\signal, t, \property_1), \qs(\signal, t, \property_2))\\
    &\qs(\signal, t, \property_1 \lor \property_2) = \max(\qs(\signal, t, \property_1), \qs(\signal, t, \property_2))\\
    &\qs(\signal, t, F_{[t_1, t_2]}\property) = \max_{t' \in [t'+t_1, t'+t_2]}\qs(\signal, t', \property)\\
    &\qs(\signal, t, G_{[t_1, t_2]}\property) = \min_{t' \in [t'+t_1, t'+t_2]}\qs(\signal, t', \property)\\
    &\qs(\signal, t, \property_1U_{[t_1, t_2]}\property_2) \\
    &= \max_{t' \in [t'+t_1, t'+t_2]}\Big(\min (\qs(\signal, t', \property_2),  \min_{t'' \in [t, t')}\qs(\signal, t'', \property_1))\Big)
        \end{split}
\end{equation}
Here, for convenience, we introduce the finally ($F$) and globally ($G$) operators which can be defined using $U$, i.e., $F_I \property = \top U_I \property$ and $G_I \property = \neg (\top U_I \neg \property)$ where $I$ is a time interval.
By the construction of Equation \eqref{eq:qs_definition}, we are able to evaluate a real-valued degree of property satisfaction of the current state so far. Therefore, people have been using this score in reward functions via various ways to guide RL \cite{hamilton2022training,li2017reinforcement,li2018policy}. However, because these metrics are derived from STL satisfaction, they still lack the expressiveness when describing certain tasks in the same way of STL. As a response, researchers have proposed sophisticated reward shaping techniques so that the score can encourage and discourage agent behaviors more accurately \cite{berducci2021hierarchical}. However, to our knowledge, little has been done on expressing ASAP, which is vital to the CPS recovery problem introduced in the next subsection.

\subsection{Real-time Cyber-physical System Recovery}
\label{subsec:cps_recovery}


Cyber-physical systems (CPS) are (usually real-time) software-controlled agents that interacts with the physical world. For instance, autonomous vehicles, robots, smart power girds and medical devices are all CPS. Recently, one problem, namely CPS recovery, have been widely discussed by researchers. Specifically, when a CPS finds itself in an undesirable state, how to guide it back to a predefined safe target set of states?

Researchers have discussed some harder versions of this problem. 
For example, some discuss the guidance when an agent's sensors are under attack so that the observed states are unreliable \cite{kong2018cyber,zhang2020real,zhang2021real}. However, most of the works assume a known environment dynamics so that the recovery controller can be synthesized by optimal control algorithms. We focus on a more severe case, where the dynamics is an unknown black-box, but we are able to sample data points from it for agent training purposes. Overall, our work explore how to patch up insufficient expressiveness of formal specifications in this real world-motivated scenario. 

%% file: sections/3_problem.tex
\section{Problem Formulation}
\label{sec:problem}

Our problem is motivated by the real-time CPS recovery problem defined in Section \ref{sec:preliminaries}. That is, in certain scenarios we want an agent, which is not currently satisfying some property, such as ASAP. For instance, in the recovery problem, we want the agent to reach a pre-defined target state set ASAP.

Specifically, we would like to build a framework that inputs a state space $\statespace$, an action space $\actionspace$, and a specified STL property $\property$. The framework outputs a control policy $\policy(\action_t | \state_t)$, i.e., a probability distribution of actions given a state, such that the satisfaction of $\property$ is ASAP.


Given a trace $\signal = \state_0\state_1\dots\state_k$ for some $k\geq 0$, we use $(\signal, t) \models \property$ where $0\leq t\leq k$ to denote that the state $\state_t$ satisfies the property $\property$. Then, we define a function $t_\property(\cdot)$ that evaluates the first time step that a signal satisfies a property as follows.
\begin{equation}\label{eq:first_sat_time}
    t_\property(\signal) = \left\{ \begin{array}{l} \infty,\qquad\qquad \nexists t \in \{0, \dots, k\}, , (\signal, t) \models \property \\
    \arg\min_{t \in \{0, \dots, k\}} ((\signal, t) \models \property),\ \ \text{otherwise.} \end{array}\right.
\end{equation}
Intuitively, $t_\property(s) = \infty$ if there is no time step that $s$ satisfies $\property$. Otherwise, it is the first time step of satisfaction.


We say that a policy $\policy^*$ is ASAP if it has a higher probability to sample a trace that satisfies the given property within a shorter time, under an unknown probabilistic underlying system model, i.e., a state transition distribution $\sysmodel(\state_{t+1} | \state_t, \action_t)$. 
\begin{definition}[ASAP Policies]
\label{def:asap}
On a state space $\statespace$, an action space $\actionspace$, and an underlying system model $\sysmodel$, we say a policy $\policy^*$ is ASAP iff for two arbitrary traces $\signal$ and $\signal'$ that both start at an arbitrary initial state $\state_0$,
\begin{equation}
    \label{eq:asap}
    t_\property(\signal) < t_\property(\signal') \implies \Pr_{\sysmodel, \policy^*}(\signal' | \state_0) < \Pr_{\sysmodel, \policy^*}(\signal | \state_0)
\end{equation}
\end{definition}

We propose an approach to find such an optimal policy based on the following assumptions.


\noindent\textbf{Assumption 1: }\textit{A system is always maintainable once property is reached.} When a system satisfies the given property $\property$, then it has the highest probability to be maintained to keep satisfying the property using some control inputs. Notice that this assumption is reasonable, since a target set in the recovery problem is often a neighborhood of a stable equilibrium~\cite{abad2016reset,abdi2018guaranteed}. Once inside this neighborhood, the agent's behavior will be stabilized and we can assume a constant expected cumulative reward in the future, i.e. for any policy $\policy$, on any trajectory $\signal'$ that continues on some $\signal$,
\begin{equation}   \label{eq:stable_equilibrium}
\small
(|\signal| = t) \land ((\signal, t) \models \property) \implies \E_\policy[\sum_{t'=t+1}^\infty \gamma^{t'} \reward(\state'_{t'}) | \signal'_{0:t} = \signal] = M
\end{equation}
where $M$ is a constant and $\signal'_{0:t}$ denotes the sub-trace from time step $0$ to $t$ of $\signal'$. In other words, if $\signal$ satisfies the property in its last time step, then every trace that extends from it has a fixed expected cumulative reward afterwards due to stability.


\noindent\textbf{Assumption 2: } \textit{The length and the semantics score of a trace are always bounded.} Firstly, our RL framework only considers the system traces which are length-bounded. We denote the bounds $k_{min}$ and $k_{max}$. Secondly, we assume that all system traces have bounded quantitative semantics scores. We denote the bounds $\qs_{min}$ and $\qs_{max}$.

\noindent\textbf{Assumption 3: } \textit{The blackbox (unknown) system performs as a Markov Decision Process.} Precisely, any action from the system is performed only based on the current state. Therefore, all traces generated by the system has a probability which is independent from the others.




We seek to solve the following two problems:\\
\noindent\textbf{Problem 1: } Given a blackbox system along with its state space $\statespace$ and action space $\actionspace$. How can we learn an ASAP policy $\policy^*$ regarding to an STL property $\property$?


\noindent\textbf{Problem 2: } How can we leverage the ASAP policy obtained by solving Problem 1 and use it to solve the system recovery problem?


%% file: sections/4_framework.tex
\section{The ASAP-Phi Framework}
\label{sec:framework}

\subsection{The Workflow}
\label{subsec:workflow}

For Problem 1, we propose the following framework, namely ASAP-Phi, which consists of generating a piece-wise reward function and training on actor-critic-based model-free RL algorithms, while the training requires sampling from the underlying transition distribution $\sysmodel(s_{t+1}|s_t, \action_t)$. 

First, we design a deterministic piece-wise reward function $\reward_{\text{asap}}$ as follows.
\begin{equation}
    \label{eq:asap_reward}
    \reward_{\text{asap}}(\signal, t, \property) =
    \begin{cases}
    \qs(\signal, t, \property), \quad\ \ \ \text{ if } (\signal, t) \not\models \property\\
    \reward_{\text{sat}} >> \qs_{\max}, \quad \text{ otherwise}
    \end{cases}
\end{equation}

Equation \eqref{eq:asap_reward} defines a reward function that evaluates on a trace $\signal$ w.r.t. an STL property $\property$ and a maximum time index $t$. It splits the reward into two cases: If the signal until $t$ does not satisfy the given property, we use the quantitative semantics score $\qs$, defined in Section \ref{subsec:spec_guided_rl}, to evaluate the degree of property satisfaction; otherwise, we pick a large constant $\reward_{\text{sat}}$ that is greater than all possible rewards in the previous case. Notice that, to compute this reward, we need to keep the whole trace $\signal$ to check the property satisfaction. This large constant will be discussed in the next subsection.

Usually, the quantitative semantics produce sparse reward since it only compute the degree of the robustness at the end of the episode which is also known as sparse reward \cite{hamilton2022training}. And the return is the degree of the robustness for the whole episode.
Alternatively, there is also a dense reward option for the quantitative semantics where the degree of the robustness is computed at each time step. The return of the dense reward represents the degree of robustness for the subtrajectory from the start to the current time step. In this paper, we proposed a variation of the dense reward: Finite dense reward which evaluates the degree of the robustness of the trajectory contains the most recent $d$ time steps, $d$ is a hyperparameter.


We present our RL framework in Algorithm~\ref{alg:asap_phi}. The main loop iteratively generates a random trace whose length is in the specific range. It progressively updates a policy (actor) model $\policy_\policyparam$ parameterized by $\policyparam$ and a value (critic) model $\criticval_\criticvalparam$ parameterized by $\criticvalparam$ by calling an external actor-critic-based algorithm, denoted by $AC$, that can be soft actor-critic (SAC) \cite{haarnoja2018soft} or deep deterministic policy gradient (DDPG) \cite{silver2014deterministic}.



\begin{algorithm}
\caption{ASAP-Phi (on-policy version)}
\label{alg:asap_phi}
\hspace*{\algorithmicindent} \textbf{Input}: Reward function $\reward_{asap}$, state space $\statespace$, action space $\actionspace$, STL property $\property$, actor-critic training algorithm $\ac$, system model $\sysmodel$ that is unknown but allows sampling, number of training samples $m$, lower and upper bound of episode length $k_{\min}$ and $k_{\max}$  \\
 \hspace*{\algorithmicindent} \textbf{Output}: Control policy $\policy_\policyparam$ 
\begin{algorithmic}[1]
\State Initialize policy model $\policy_\policyparam$ and value model $\criticval_\criticvalparam$
\State $m_{train} \gets 0$
\While{$m_{train} < m$}
    \State $k \gets $ uniformly sample from $[k_{\min}, k_{\max}]$
    \State $\state_0 \gets$ uniformly sample from $\statespace$
    \State Cached signal $\signal \gets [\state_0]$
    \For{$t = 0, \dots, k$}
        \State $\reward \gets \reward_{\text{asap}}(\signal, t,  \property)$
        \State $\action_t \gets$ sample from $\policy_\policyparam(\cdot|\state_t)$
        \State $\state_{t+1} \gets $ sample from $\sysmodel(\cdot|\state_t, \action_t)$
        \State Append $\state_{t+1}$ to $\signal$
        \State $\policy_\policyparam$, $\criticval_\criticvalparam \gets AC(\state_t, \action_t, \state_{t+1}, \reward)$
        \State $m_{train} \gets m_{train} + 1$
    \EndFor
\EndWhile
\end{algorithmic}
\end{algorithm}

Algorithm \ref{alg:asap_phi} trains the policy and value models on $m$ data points, which combines to multiple trace episodes of lengths between $k_{\min}$ and $k_{\max}$. Specifically, the outer loop from line 3 to 14 represents the entire procedure on all episodes, and the inner loop from line 7 to 13 is one episode. The training is at line 11, where the actor-critic algorithm updates the two models using $\signal_t$, $\action_t$ and $\signal_{t+1}$, on our designed reward function $\reward_{\text{asap}}$. 
Notice that this version of algorithm is on-policy. That is, there is one model update per arrival of new data point. We can also leverage off-policy training by moving the model update at the end of every episode.

\subsection{Proof of Correctness}
\label{subsec:analysis}

In this subsection, we prove the correctness of ASAP-Phi, i.e., if ASAP-Phi returns an optimal policy, then it is also ASAP.


\begin{theorem}[ASAP-Phi Correctness]
\label{thm:asap_phi_correctness}
The optimal policy learned from ASAP-Phi framework is an ASAP policy.
\end{theorem}

\begin{figure}
    \centering
    \includegraphics[width=0.5\textwidth]{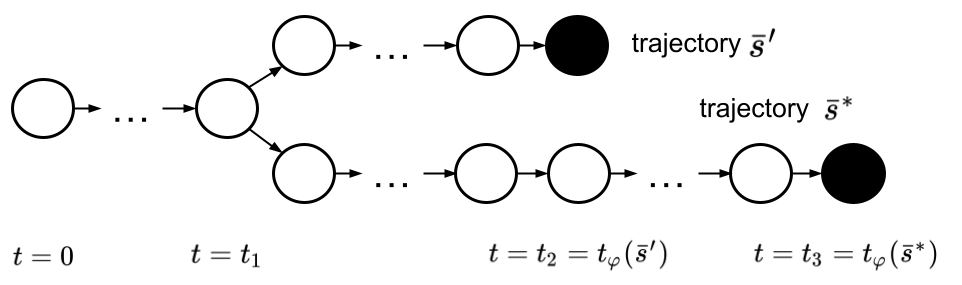}
    \caption{The two existing trajectories under the assumed $\policy^*$. An empty circle represents the trajectory so far does not satisfy $\property$, while a filled circle means satisfy.}
    \label{fig:proof_trajectories}
\end{figure}

\begin{proof}
We prove the theorem by contradiction. Assume that there is an optimal policy $\policy^*$ learned by ASAP-Phi, and it is not ASAP. We seek to prove that $\policy^*$ is also not optimal that contradicts our assumption.

Given that $\policy^*$ is optimal but not ASAP, then there exists two traces $\bar{s}^*$ and $\bar{s}'$ both of which start from an initial state $s_0$ and reach a target state as their last states. We also have that $\Pr_{\sysmodel, \policy^*}(\signal'| \state_0) < \Pr_{\sysmodel, \policy^*}(\signal^* | \state_0)$ and $|\bar{s}'| < |\bar{s}^*|$ by assumption, i.e., $\bar{s}^*$ is longer than $\bar{s}'$ but the former one has a higher probability to be performed than the later one using $\policy^*$. We illustrate these two traces in Figure~\ref{fig:proof_trajectories}. Here, $t_1$ is the last time that the two traces are exactly identical since beginning, $t_2$ is the time for $\signal'$ to satisfy $\property$ for the first time, and $t_3$ is for $\signal^*$.


By Assumption 1, a trace can be maintained in the target set when it reaches there. We therefore construct a trajectory by padding satisfactory states after $\signal'$ until time $t_3$. We denote this trajectory as $\signal''$. Then we have that, $\Pr_{\sysmodel, \policy^*}(\signal''|\state_0) \leq
    \Pr_{\sysmodel, \policy^*}(\signal'|\state_0) < \Pr_{\sysmodel, \policy^*}(\signal^*|\state_0)$, since $\bar{s}''$ is an extension of $\bar{s}'$.

Since the $\criticval^*$ is also optimal, it is a solution to the Bellman equation \eqref{eq:ac_bellman}:
\begin{equation}
    \label{eq:split_1}
    \small
    \begin{split}
        \criticval^*_{\policy}(\state_0) &= \E_\policy[\sum_{t=0}^\infty \gamma^t \reward_{\text{asap}}(\signal, t, \property) | \state_0]
        = \E_\policy[\sum_{t=0}^{t_3} \gamma^t \reward_{\text{asap}}(\signal, t, \property) | \state_0] \\
        &+\underbrace{\E_\policy[\sum_{t=t_3+1}^\infty \gamma^t \reward_{\text{asap}}(\signal, t, \property) | \state_0]}_{\text{denote this term $D_\policy$}}
    \end{split}
\end{equation}
and therefore
\begin{equation}
    \label{eq:split_2}
 \small
    \begin{split}
        &\criticval^*_{\policy^*}(\state_0) = \E_{\policy^*}[\sum_{t=0}^{t_3} \gamma^t \reward_{\text{asap}}(\signal, t, \property) | \state_0]
        + D_{\policy^*}\\
        & = p_1\underbrace{\sum_{t=0}^{t_3} \gamma^t \reward_{\text{asap}}(\signal^*, t, \property)}_{\text{denote this term $A$}}  + p_2\underbrace{\sum_{t=0}^{t_3} \gamma^t \reward_{\text{asap}}(\signal'', t, \property)}_{\text{denote this term $B$}}\\
        & + \underbrace{\E_{\policy^*}[\sum_{t=0}^{t_3} \gamma^t \reward_{\text{asap}}(\signal, t, \property) | \state_0, \signal \neq \signal^* \land \signal \neq \signal'' \land |\signal| = t_3)]}_{\text{denote this term $C_{\policy^*}$}} + D_{\policy^*}\\
        & = p_1A + p_2B + C_{\policy^*} + D_{\policy^*}
    \end{split}
\end{equation}
where $p_1 = \Pr_{\sysmodel, \policy^*}(\signal^*|\state_0)$, $p_2 = \Pr_{\sysmodel, \policy^*}(\signal''|\state_0)$. That is, the probabilities to take traces $\signal^*$ and $\signal''$, respectively. The term $C_{\policy^*}$ denotes the value to be obtained by taking any other trace of length $t_3$ starting from $\state_0$ under policy $\policy^*$.

We now construct an alternative policy $\policy'$, by swapping the probabilities $p_1$ and $p_2$ on the two traces. We are able to do so because of Assumption 3, that all trajectories are independent from each other due to the Markovian state transitions. Then we have the following new value estimation:
\begin{equation}
\label{eq:alternative_value}
\criticval^*_{\policy'}(\state_0) = p_2A + p_1B + C_{\policy'} + D_{\policy'}
\end{equation}
Since all the traces of the length $t_3$ other than $\bar{s}^*$ and $\bar{s}''$ have the same probabilities in $\policy^*$ and $\policy'$, thus $C_{\policy'} = C_{\policy^*}$. Moreover, by Assumption 1, since both $\signal''$ and $\signal^*$ enters the stable neighborhood at time $t_3$, we can leverage Equation \eqref{eq:stable_equilibrium} and get
\begin{equation}
\small
    \begin{split}
        &D_{\policy'} = \E_\policy[\sum_{t=t_3+1}^\infty \gamma^t \reward_{\text{asap}}(\signal, t, \property) | \state_0]\\
        &= \E_\policy[\sum_{t=t_3+1}^\infty \gamma^t \reward_{\text{asap}}(\signal, t, \property) | \state_0, \signal_{0 : t_3} = \signal^*] \\
        &+ \E_\policy[\sum_{t=t_3+1}^\infty \gamma^t \reward_{\text{asap}}(\signal, t, \property) | \state_0, \signal_{0 : t_3} = \signal''] \\
        &+ \E_\policy[\sum_{t=t_3+1}^\infty \gamma^t \reward_{\text{asap}}(\signal, t, \property) | \state_0, \signal \neq \signal^* \land \signal \neq \signal'']\\
        &= 2M + \E_\policy[\sum_{t=t_3+1}^\infty \gamma^t \reward_{\text{asap}}(\signal, t, \property) | \state_0, \signal_{0 : t_3} \neq \signal^* \land \signal_{0 : t_3} \neq \signal'']\\
        & = D_{\policy^*}
    \end{split}
\end{equation}
where $\signal_{0 : t_3}$ denotes the sub-trace from time step $0$ to $t_3$ of a trace $\signal$. Therefore, we also have that $D_{\policy'} = D_{\policy^*}$. Then we only need to compare $p_1A + p_2B$ and $p_2A + p_2B$.

By the definition of $\reward_{\text{asap}}$ in \eqref{eq:asap_reward}, we have that
\begin{equation}
 \small
    \begin{split}
        A &= \underbrace{\sum_{t=0}^{t_1}\gamma^t\qs(\signal^*, t, \property)}_{A_1} + \underbrace{\sum_{t=t_1+1}^{t_2}\gamma^t\qs(\signal^*, t, \property)}_{A_2} + \underbrace{\sum_{t=t_2+1}^{t_3}\gamma^t\qs(\signal^*, t, \property)}_{A_3}\\
        B &= \underbrace{\sum_{t=0}^{t_1}\gamma^t\qs(\signal'', t, \property)}_{B_1} + \underbrace{\sum_{t=t_1+1}^{t_2}\gamma^t\qs(\signal'', t, \property)}_{B_2} + \underbrace{\sum_{t=t_2+1}^{t_3}\gamma^t \reward_{\text{sat}}}_{B_3}
    \end{split}
\end{equation}
Since the two traces $\signal^*$ and $\signal''$ are identical from $t=1$ to $t=t_1$, we have that $A_1 = B_1$. Also, since $\reward_{\text{sat}} > \qs_{\max}$, we have that $A_3 < B_3$. For the last pair $A_2$ and $B_2$, we have that
\begin{equation}
\small
    \begin{split}
        A_2 - B_2 &= \sum_{t=t_1+1}^{t_2}\gamma_t(\qs(\signal^*, t, \property) - \qs(\signal'', t, \property))\\
        &\leq \sum_{t=t_1+1}^{t_2}\gamma_t(\qs_{\max} - \qs_{\min})
        \leq (\qs_{\max} - \qs_{\min})\sum_{t=0}^{k_{\max}}\gamma_t\\
        &\leq (\qs_{\max} - \qs_{\min})k_{\max}
    \end{split}
\end{equation}
Therefore, if we set $r_{\text{sat}} > \qs_{max} + (\qs_{\max} - \qs_{\min})k_{\max}$, we can guarantee that $A_2 - B_2 < B_3 - A_3$. Hence, we have $A < B$. Recall that $p_1 > p_2$, so $p_2A + p_1 B > p_1A + p_2B$, i.e., $\criticval^*_{\policy'}(\state_0) > \criticval^*_{\policy^*}(\state_0)$. Therefore, $\policy'$ is better than $\policy^*$, or $\policy^*$ is not an optimal policy.
\end{proof}

\subsection{ASAP-Phi in Recovery Problem}
\label{subsec:asap_recovery}
ASAP-Phi is a general solution that can drive a system from a deviated state (that does not satisfy a property) to a target state (that satisfies it).
While not explicitly dedicated to the security aspect, our solution positively enhances the resilience to faults and attacks.
For specific fault or attack scenarios, certain adaptations to our solution may be needed.
As an use case, we propose the following use of ASAP-Phi to solve the CPS recovery problem and answer Problem 2.

In the recovery problem, we have a target set of states $\statespace_T \subset \statespace$ that we want an agent to reach it ASAP. Optionally, there may exist certain states $\statespace_O \subset \statespace$ that represents obstacles or danger zones that we want to avoid, $\statespace_T \cap \statespace_0 = \emptyset$. The target set and obstacle set are assumed to be bounded, so We can check whether a state $\state_t$ is contained inside efficiently. Therefore, we specify these two requirements using STL properties:
\begin{equation}
    \label{eq:recovery_properties}
        \property_T = \state_t \in \statespace_T, \quad
        \property_O = \state_t \not\in \statespace_O
\end{equation}

In other words, we want to learn a policy that satisfies $\property_T$ ASAP while always satisfying $\property_O$. Therefore, we need to measure the degree of robustness for $\property_T$ as reward for training. Additionally, we need to have reward shaping to make sure the optimal policy can satisfy $\property_O$. We leverage the ASAP-Phi training in Algorithm \ref{alg:asap_phi} to do so, by taking regular policy and value function updates on $\property_T$, and early stop once a sampled trace violates $\property_O$. For convenience, we wrap up the initialization of ASAP-Phi for each episode as a canonical subroutine $ASAP\_phi\_ep\_init()$ (line 4-5 of Algorithm \ref{alg:asap_phi}), and each update step in the inner loop as $ASAL\_phi\_step(\property)$ (line 8-12 of Algorithm \ref{alg:asap_phi}), with the reward function designed based on a property $\property$. The solution is stated in Algorithm \ref{alg:asap_phi_recovery}

\begin{algorithm}
\caption{ASAP-Phi for Recovery}
\label{alg:asap_phi_recovery}
\hspace*{\algorithmicindent} \textbf{Input}: Properties $\property_T$, $\property_O$ and everything else required by Algorithm \ref{alg:asap_phi} \\
 \hspace*{\algorithmicindent} \textbf{Output}: Recovery control policy $\policy_\policyparam$ 
\begin{algorithmic}[1]
\State Initialize policy model $\policy_\policyparam$ and value model $\criticval_\criticvalparam$
\State $m_{train} \gets 0$
\While{$m_{train} < m$}
\State $ASAP\_phi\_ep\_init()$
    \For{$t = 0, \dots, |\signal|-1$}
    \If{$(\signal, t) \not\models \property_O$}
        \State break
    \EndIf
    \State $\policy_\policyparam$, $\criticval_\criticvalparam \gets ASAL\_phi\_step(\property_T)$
    \State $m_{train} \gets m_{train} + 1$
    \EndFor
\EndWhile
\end{algorithmic}
\end{algorithm}

Notice that the key idea is to search for an optimal policy that is ASAP to satisfy $\property_T$ and early stop on violation of $\property_O$. The reason to do so is to prevent the trajectory from getting the high reward $\reward_{sat}$ if it enters an obstacle state, such that the agent will learn to take alternative paths to reach a target state ASAP. We  demonstrate the effectiveness of this approach in the experiment section.

%% file: sections/5_experiments.tex
\section{Experimental Evaluations}
\label{sec:experiments}

\subsection{Setup}
\label{subsec:expr_setup}


\begin{table*}[h]
\centering
\begin{tabular}{cccccccccc}
\toprule
  \multicolumn{1}{c}{Max} &  \multicolumn{1}{c}{\multirow{2}{*}{Method}} & \multicolumn{2}{c}{DC Motor Position}& \multicolumn{2}{c}{Bicycle} & \multicolumn{2}{c}{Attitude Control}& \multicolumn{2}{c}{Swimmer}\\ 
\cmidrule{3-10}
  Time Tolerance*&  & \textit{R}  & \textit{R\&A} & \multicolumn{1}{c}{\textit{R}}  & \textit{R\&A} & \textit{R}  & \textit{R\&A} & \multicolumn{1}{c}{\textit{R}}  & \textit{R\&A} \\ 
\midrule
\multicolumn{1}{c}{\multirow{3}{*}{30}} & ASAP-phi & \multicolumn{1}{c}{\textbf{97.4\%}} & \textbf{97.8\%} & \multicolumn{1}{c}{\textbf{92.8\%}} & \textbf{91.0\%} & \multicolumn{1}{c}{\textbf{93.9\%}} & \textbf{92.0\%}       & \multicolumn{1}{c}{\textbf{62.6\%}} & \textbf{64.1\%}  \\ 
\cmidrule{2-10} 
\multicolumn{1}{c}{} & $\qs$-reward & \multicolumn{1}{c}{83.5\%} & 82.1\%  & \multicolumn{1}{c}{80.6\%}  & 86.4\% & \multicolumn{1}{c}{75.1\%} & 73.6\% & \multicolumn{1}{c}{58.4\%} & 57.2\%  \\ 
\cmidrule{2-10}
\multicolumn{1}{c}{} & $d$-reward & \multicolumn{1}{c}{35.5\%}  & 95.3\% & \multicolumn{1}{c}{78.4\%}& 27.6\%  & \multicolumn{1}{c}{21.1\%} & 12.7\% & \multicolumn{1}{c}{42.5\%} & 6.5\%\\ 
\midrule %
\multicolumn{1}{c}{\multirow{3}{*}{25}} & ASAP-phi & \multicolumn{1}{c}{\textbf{95.2\%}} & \textbf{95.3\%} & \multicolumn{1}{c}{\textbf{88.1\%}} & \textbf{89.2\%} & \multicolumn{1}{c}{\textbf{86.5\%}} & \textbf{85.3\%} & \multicolumn{1}{c}{\textbf{57.4\%}} & \textbf{59.3\%}       \\ 
\cmidrule{2-10} 
\multicolumn{1}{c}{} & $\qs$-reward & \multicolumn{1}{c}{77.3\%} & 79.8\% & \multicolumn{1}{c}{{80.6\%}} & 86.1\% & \multicolumn{1}{c}{72.4\%} & 70.9\% & \multicolumn{1}{c}{51.5\%} & 50.8\%  \\
\cmidrule{2-10} 
\multicolumn{1}{c}{} & $d$-reward & \multicolumn{1}{c}{32.2\%} & 95.1\% & \multicolumn{1}{c}{76.2\%} & 27.0\%  & \multicolumn{1}{c}{18.0\%} & 11.9\% & \multicolumn{1}{c}{39.4\%}  & 6.1\%  \\ 
\midrule
\multicolumn{1}{c}{\multirow{3}{*}{15}} & ASAP-phi & \multicolumn{1}{c}{\textbf{89.0\%}} & 84.9\%   & \multicolumn{1}{c}{\textbf{74.4\%}}  & 65.1\% & \multicolumn{1}{c}{\textbf{55.7\%}}          & \textbf{52.0\%}             & \multicolumn{1}{c}{\textbf{52.0\%}} & \textbf{53.0\%}  \\ 
\cmidrule{2-10} 
\multicolumn{1}{c}{} & $\qs$-reward  & \multicolumn{1}{c}{59.2\%} & 79.4\% & \multicolumn{1}{c}{72.6\%} & \textbf{65.5\%} & \multicolumn{1}{c}{52.2\%} & 51.9\%       & \multicolumn{1}{c}{47.4\%}          & 46.0\% \\ 
\cmidrule{2-10} 
\multicolumn{1}{c}{} & $d$-reward & \multicolumn{1}{c}{29.2\%}  & \textbf{87.0\%}  & \multicolumn{1}{c}{65.4\%}  & 25.3\%  & \multicolumn{1}{c}{15.6\%}          & 10.7\%                & \multicolumn{1}{c}{34.1\%}          & 4.8\% \\
\bottomrule
\end{tabular}
\caption{Agent performance measured by successful recovery rate. Here, $R$ and $R\&A$ denote $Reach$ and $Reach\&Avoid$ tasks, respectively. *: For swimmer, the maximum time tolerance are 50, 40, 30 time steps instead of 30, 25, 15.}
\label{tab:dcbicycle}
\end{table*}

We set up groups of experiments on recovery problems for evaluation. The experiments are implemented on four benchmarks: DC Motor Position, Bicycle, Attitude Control and Mujoco Swimmer. DC Motor Position is a classical system from control community and CPS community, with a goal to control a motor's rotation angle by direct current \cite{zhang2020real}. Bicycle benchmark simulates the dynamics of a bicycle with two control inputs: steering angle and acceleration \cite{kong2015kinematic}, with a goal to ride the bicycle at different speeds and directions. Attitude Control benchmark is from the artificial intelligence category of Workshop on Applied Verification for Continuous and Hybrid Systems (ARCH-COMP'22) \cite{lopez2022arch}, with a goal to control the rigid body to avoid unsafe area. Finally, the Mujoco Swimmer benchmark is one of the well-known Mujoco environments for reinforcement learning \cite{brockman2016openai,todorov2012mujoco}, aiming to control the two-joints system and move to the right as fast as possible. More detailed descriptions of the benchmarks can be found in Appendix. Although some of the benchmarks have formal models such as ODEs, our approach only uses their simulators as black boxes.

We select classical actor-critic based training algorithms including SAC, DDPG, A2C, TD3 and PPO. To avoid the performance decay from the implementation of these algorithms, we use the implementation from stable baselines 3 which developed and maintained by OpenAI \cite{stable-baselines3}. For each benchmark, there is a given target set and a unsafe set. These two sets are defined as balls for simplification. We train agents to find policies that make the system reach the target set ASAP without touching the unsafe set.

\begin{table*}[h]
\centering
\begin{tabular}{ccccccccc}
\toprule
                           & \multicolumn{2}{c}{DC Motor Position}                       & \multicolumn{2}{c}{Bicycle}                                 & \multicolumn{2}{c}{Attitude Control}                        & \multicolumn{2}{c}{Swimmer}                                 \\ \cmidrule{2-9} 
                           & \multicolumn{1}{c}{\textit{Reach}}  & \textit{Reach\&Avoid} & \multicolumn{1}{c}{\textit{Reach}}  & \textit{Reach\&Avoid} & \multicolumn{1}{c}{\textit{Reach}}  & \textit{Reach\&Avoid} & \multicolumn{1}{c}{\textit{Reach}}  & \textit{Reach\&Avoid} \\ \midrule
\multicolumn{1}{c}{SAC}  & \multicolumn{1}{c}{97.4\%}          & \textbf{97.8\%}       & \multicolumn{1}{c}{\textbf{83.0\%}} & 78.8\%                & \multicolumn{1}{c}{75.6\%}          & 74.1\%                & \multicolumn{1}{c}{62.6\%}          & \textbf{64.1\%}       \\ \midrule
\multicolumn{1}{c}{DDPG} & \multicolumn{1}{c}{\textbf{99.6\%}} & 95.7\%                & \multicolumn{1}{c}{75.8\%}          & 78.2\%                & \multicolumn{1}{c}{62.7\%}          & 61.6\%                & \multicolumn{1}{c}{11.5\%}          & 11.8\%                \\ \midrule
\multicolumn{1}{c}{A2C}  & \multicolumn{1}{c}{56.8\%}          & 39.3\%                & \multicolumn{1}{c}{37.8\%}          & 22.8\%                & \multicolumn{1}{c}{15.2\%}          & 14.3\%                & \multicolumn{1}{c}{0.6\%}           & 0.4\%                 \\ \midrule
\multicolumn{1}{c}{TD3}  & \multicolumn{1}{c}{99.0\%}          & 96.3\%                & \multicolumn{1}{c}{76.8\%}          & \textbf{81.4\%}       & \multicolumn{1}{c}{\textbf{78.4\%}} & \textbf{77.6\%}       & \multicolumn{1}{c}{\textbf{62.8\%}} & 54.1\%                \\ \midrule
\multicolumn{1}{c}{PPO}  & \multicolumn{1}{c}{37.8\%}          & 39.6\%                & \multicolumn{1}{c}{79.9\%}          & 80.2\%                & \multicolumn{1}{c}{65.0\%}          & 64.8\%                & \multicolumn{1}{c}{15.2\%}          & 13.1\%                \\ \bottomrule
\end{tabular}
\caption{Success rate of traces produced by policies trained by different algorithms, the maximum time tolerance is set to 30 time steps for DC Motor Position, 20 time steps for Bicycle and Attitude Control, the maximum time tolerance is set to 50 for Swimmer benchmark.} 
\label{tab:algoperformance}
\end{table*}

Despite our proof in Section \ref{subsec:analysis}, it is generally hard to find an optimal policy by training, so we seek sub-optimal policies, i.e., recovery within a selected maximum time tolerance, which is similar to the evaluation metrics for recovery tasks used in previous work \cite{zhang2020real,zhang2021real,kong2018cyber}.
We consider two types of recovery tasks: $Reach$ and $Reach\&Avoid$. For $Reach$ tasks, unsafe obstacle states are absent. For $Reach\&Avoid$ tasks, the system should avoid a pre-defined obstacle set, as described in Section \ref{subsec:asap_recovery}. We test the trained agents on 1000 random initial points and count how many traces can reach the target set within the maximum time tolerance. A trace will be terminated if it touches the obstacle and considered a failure for $Reach\&Avoid$ tasks. We calculate the success rate of the traces generated by policies to evaluate their performance. 

We compare the performance between ASAP-Phi and other baseline rewards as main results. Additionally, we do ablation studies on ASAP-Phi to show ASAP-Phi's insensitivity to training algorithms. For more details about experiment, please see appendix.


\subsection{Main Results}
\label{subsec:expr_main_results}
We choose two reward function baselines to compare with ASAP-Phi. One is quantitative semantics reintroduced in section \ref{subsec:spec_guided_rl}, which represents the degree of satisfaction of a given STL specification \cite{hamilton2022training}. We denote this reward as $\qs$-reward. Another reward function baseline is heuristic distance-based reward function which is widely applied in existed works \cite{burchfiel2016distance,trott2019keeping,li2017reinforcement}, and we restate it as follows to avoid abouse of notations.
\begin{equation}
    \reward_d = -\lambda_Td_T+\lambda_Od_O + \reward_{base}
\end{equation}
where $d_T$ and $d_O$ are the distances to the target set $\statespace_T$ and the unsafe obstacle set $\statespace_O$, respectively, and $\lambda_T$, $\lambda_O$ and $\reward_{base}$ are hyperparameters.

We use SAC to train the agents for the main results. From Table \ref{tab:dcbicycle}, there are three main observations:
\begin{enumerate}
    \item ASAP-Phi outperform baselines in most cases. We can see ASAP-Phi have more than $50\%$ success rate for every cases and defeats the two baselines for most trials. In only a few cases, for example, the $Reach\&Avoid$ task on bicycle benchmark, the $\qs$-reward performs slightly better than ASAP-Phi. This small difference can be from due to the noise in training since there is no guarantee that the optimal policy is found.
    \item ASAP-Phi's performance is much more stable than the other two baselines. For example, the $d$-reward baseline performs much worse than others on most tasks but slightly outperform others on $Reach\&Avoid$ task for DC Motor Position if the maximum time tolerance is 15 time steps. This instability does not happen to ASAP-Phi.
    \item Sometimes an agent can achieve better performance on $Reach\&Avoid$ tasks than $Reach$ tasks. Intuitively, $Reach\&Avoid$ tasks are harder than $Reach$ tasks. However, it is hard to find optimal policies through training, so there can be performance difference between tasks.
\end{enumerate}
Since ASAP-Phi outperforms the baselines for the majority of cases, we state that ASAP-Phi is effective to achieve a given specification ASAP in practice.

\begin{figure}
    \centering
    \centering
    \includegraphics[width=0.45\textwidth] {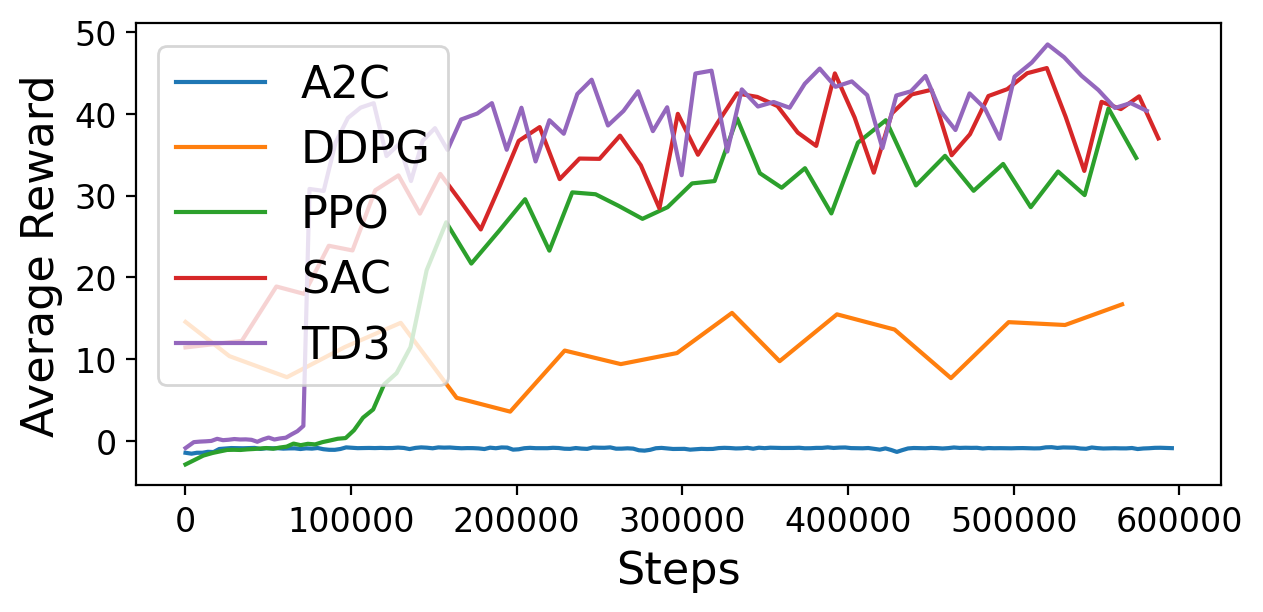}
    \caption{Training history of Swimmer on $Reach\&Avoid$ task}
\label{fig:training}
\end{figure}

\subsection{Ablation Studies}
\label{subsec:expr_ablation}
We conduct ablation studies on (1) different actor-critic algorithms and (2) different ways to encourage ASAP other than the current reward functions. Here, we present the results of the former in Table \ref{tab:algoperformance}, with two main observations.
\begin{enumerate}
    \item ASAP-Phi is not designed for a particular training algorithm. In other words, Table 2 have shown the insensitivity to the training algorithms. On contrary, ASAP-Phi is compatible with various actor-critic-based methods. For example, SAC and TD3 perform the best among the five algorithms tested. DDPG also have good performance on several tasks as Fig.~\ref{fig:training} shows.
    \item The performance of PPO and A2C are not stable. For example, PPO perform poorly on the Swimmer benchmark and A2C perform worse than other algorithms in general for the two tasks.
\end{enumerate}

In the second ablation study on different ways of encouraging ASAP, we found that our current reward performs the best. Please refer to Appendix for the results and other details.

%% file: sections/6_discussion.tex
\section{Discussion}
\label{sec:discussion}

\textbf{Advantages of ASAP-Phi.} From the analysis in Section \ref{subsec:analysis} we prove that ASAP-Phi is seeking an optimal policy that maximizes the probability to choose a trace to satisfy a given formal specification ASAP, using only sampled training data but no system dynamics knowledge. To our knowledge, such a framework is yet to be proposed by existing work, despite its necessity in various applied scenarios such as CPS recovery. Moreover, although it is commonly hard to achieve this theoretical optimal policy in implementation, our experiments in Section \ref{sec:experiments} show that ASAP-Phi achieve high success rate to recover an agent within deadline, up to $97\%$ and defeating state-of-the-art baselines in general.

\noindent\textbf{Limitations of ASAP-Phi.} However, there indeed exist certain limitations. First, ASAP-Phi is based on model-free RL, which means it naturally inherits the disadvantages of this class of learning algorithms. For example, it requires collection of sufficient training data and tuning of learning hyperparameters. Fortunately, this is partially resolved by researchers designing stable actor-critic backbones \cite{stable-baselines3}. Second, although we have explored various ways to encourage ASAP, the reward function \ref{eq:asap_reward} may not necessarily be the best approach. Therefore, how to encourage the learning of an ASAP policy remains an open problem. Finally, the framework makes several assumptions. Although Assumptions 1-3 in Section \ref{sec:problem} are reasonable in the current context, so far it is unknown how to lift these assumptions while still obtaining an ASAP policy. The system dynamics, though a black box, is assumed to be static and allowing an arbitrary amount of data sampling - yet another assumption to be lifted. We believe these can be topics studied in future works.


%% file: sections/7_conclusion.tex
\section{Conclusion}
\label{sec:conclusion}

Motivated by the real-time CPS recovery problem under unknown system dynamics, we propose ASAP-Phi, a model-free RL framework that encourages fulfilling an STL formal specification ASAP. We prove that the optimal policy learned by ASAP-Phi is indeed ASAP and apply it back to the motivating case study of recovery. Experiments show that control policies trained by ASAP-Phi constantly achieve high success rate of recovery within the maximum time tolerance and generally defeat baselines on state-of-the-art benchmarks.

%% file: sections/8_appendix.tex
\begin{appendices}

\section{Detailed Experiment Setup}
\label{sec:detailed_experiment_setup}
In this section, we show and explain the details of the experiments setup including benchmarks information and hyper-parameters of the framework.

\subsection{Benchmarks}
\label{subsec:benchmarks}
We have implemented our framework on four benchmarks as Table \ref{tab:benchmark} shows. The four benchmarks have different numbers of dimensions of state space from 3 to 10, and numbers of dimensions of action space from 1 to 3. All of them are continuous systems as required by the training algorithms. All of them are non-linear systems except the DC Motor Position benchmarks.

The states of the DC Motor Position benchmark are the rotation angle, angular velocity and the current. The only control input for this benchmark is the voltage.
The states of the Bicycle benchmark are the position coordinates, the orientation and the velocity. The control inputs of the bicycle benchmark are the steering angle and acceleration. The states for the Attitude Control benchmarks are the angular velocities from three dimensions and the rodrigues parameters. The control inputs for the attitude control benchmark are the torques. The states of the Swimmer benchmarks are the position coordinates, angle of the front tip, angle of the rotors, velocities of the tip, angular velocity of the tip and the angular velocities of the rotors. The control inputs are the torques of the two rotors.

The Swimmer benchmark is the most complex benchmark among the four benchmarks. The swimmer agent consists of at least three segments in a row, with every two consecutive segments connected by an articulation joint. The swimmer stays inside a two-dimensional pool and the objective is to move as quickly as possible to the right by applying torque to the rotors and utilizing fluid friction.


\subsection{Hyper-parameters}
\label{subsec:hyper_params}

\begin{table}[!h]
\centering
\begin{tabular}{|c|c|c|}
\hline
& A2C  & PPO  \\ \hline
learning rate  & 7e-4   & 3e-4  \\ \hline 
update frequency & every 5 steps & every 2048 steps \\ \hline
gamma  & 0.99  & 0.99  \\ \hline
value function coef & 0.5 & 0.5 \\ \hline
RMSProp epsilon  & 1e-5 & NA \\ \hline
max gradient norm  & 0.5  & 0.5  \\ \hline
\end{tabular}
\caption{Hyper-parameters of A2C and PPO}
\label{tab:h1}
\end{table}

\begin{table}[!h]
\centering
\begin{tabular}{|c|c|c|c|}
\hline                                                                                                              & DDPG & TD3  & SAC  \\ \hline
learning rate          & 1e-3 & 3e-4 & 3e-4 \\ \hline
batch size              & 100  & 100  & 256  \\ \hline
buffer size             & 1e6  & 1e6  & 1e6  \\ \hline
gamma                   & 0.99 & 0.99 & 0.99 \\ \hline
soft update coefficient & 5e-3 & 5e-3 & 5e-3 \\ \hline
\end{tabular}
\caption{Hyper-parameters of DDPG,TD3 and SAC}
\label{tab:h2}
\end{table}


There are several hyper-parameters in our algorithm including training and testing. For all trials, we have the following two hyper-parameters.

(1) Maximum trace length during training. This hyper-parameter is set to 30 for all benchmarks except swimmer which is set to 50, and it remains the same as the maximum deadline in the testing phase. It is significant to choose an appropriate maximum trace length for training. On one hand, if it is too short, the system may not have enough time to reach the target set based on physical laws. If the system can never touch the target set during training, our framework will degenerate and cannot get encouragement from ASAP. On the other hand, if the maximum trace length is too long, it takes more steps for training and becomes time-consuming. Assuming the number of training steps is fixed, the longer maximum trace length is, the fewer number of traces will be explored. Insufficient exploration can be a reason for performance decay, since each trace starts from a random initial point in state space, and less proportion of the state space is explored since less number of traces are experienced. If this parameter is set to larger values such as 40 or 60, the agents can achieve a similar trend with more training steps.

(2) Total number of training steps. This hyper-parameter is set to 1e6 for DC Motor Position benchmarks, 3e6 for bicycle benchmark and Attitude Control benchmark, and 6e6 for Swimmer benchmarks. Intuitively, the more complex the system dynamics are, the more training steps are required for convergence. More evidence can be found in Appendix \ref{sec:detailed_experiment_results}.

We then list some important hyper-parameters specific to the training algorithms of our implementation in Table \ref{tab:h1} and \ref{tab:h2}. Specifically, Table \ref{tab:h1} shows the hyper-parameters for A2C and PPO algorithm, explained as follows.
\begin{enumerate}
    \item The value function coef stands for the value function coefficient for the loss calculation.
    \item RMSProp is a variation of the AdaGrad method and achieves good performance on various problems especially non-convex optimization. The RMSProp Epsilon stabilizes square root computation in RMSProp update denominator.
    \item The maximum gradient norm control the gradient clip range.
\end{enumerate}
Aside from the above, the PPO algorithm has a factor 0.95 for trade-off between bias and variance for generalized advantage estimator. And for every 10 epochs, the surrogate loss will be optimized again.

Moreover, Table \ref{tab:h2} shows the details of DDPG, TD3 and SAC algorithms of our experiments, explained as follows.
\begin{enumerate}
    \item The learning rate of DDPG is greater than other algorithms. This setting is general for training.
    \item The buffer size is greater than the training steps for our experiments. In other words, we don't pop out samples from the replay buffer during the training.
\end{enumerate}
Aside from above, the networks for all algorithms are multi-layer perceptrons.


\subsection{Tasks details}
\begin{figure}[!h]
    \centering
    \begin{subfigure}{0.4\textwidth}
    \centering
    \includegraphics[width=0.9\textwidth] {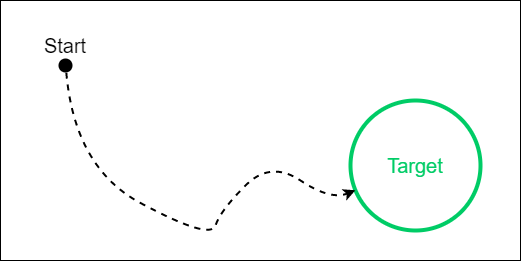}
    \caption{$Reach$ task}
    \end{subfigure}
    \begin{subfigure}{0.4\textwidth}
    \centering
    \includegraphics[width=0.9\textwidth] {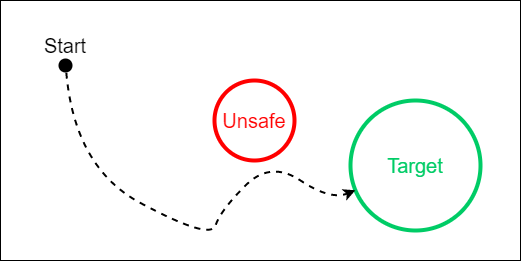}
    \caption{$Reach\&Avoid$ task}
    \end{subfigure}
    \caption{Task examples}
    \label{fig:tasks}
\end{figure}

We consider two kinds of tasks in our experiments: $Reach$ task and $Reach\&Avoid$ task as Fig.~\ref{fig:tasks} shows. We do not make any assumption of the initial state.

\begin{table}[!h]
\centering
\begin{tabular}{|c|c|c|}
\hline
              & DC Motor Position & Bicycle                    \\ \hline
\# state dims   & 3                 & 4                         \\ \hline
\# action dims  & 1                 & 2                          \\ \hline
linearity     & linear            & non-linear                 \\ \hline
continuity    & continuous        & continuous                 \\ \hline
target center & {[}$\pi$/2, 0, 0{]}   & {[}1,1,0, $\sqrt{2}${]}       \\ \hline
target radius & 0.5               & 0.8                        \\ \hline
unsafe center & {[}$\pi$/4, 0, 0{]}   & {[}0.5,0.5,0, $\sqrt{2}/2${]} \\ \hline
unsafe radius & 0.2               & 0.3                       \\ \hline \hline
& Attitude Control     & Swimmer  \\ \hline
\# state dims & 6                   & 10   \\ \hline
 \# action dims & 3                   & 2     \\ \hline linearity & non-linear          & non-linear                  \\ \hline
 continuity & continuous          & continuous  \\ \hline
 target center & {[}0,0,0,0,0,0{]}   & {[}0,9,0,0,0,0,0,0,0,0,0{]} \\ \hline
 target radius & 0.8                 & 0.8                         \\ \hline
 unsafe center & {[}0,0,0.2,0,0,0{]} & {[}0,0.6,0,0,0,0,0,0,0,0{]} \\ \hline
  unsafe radius & 0.3                 & 0.3                         \\ \hline
 
\end{tabular}
\caption{Benchmark Details}
\label{tab:benchmark}
\end{table}

\begin{figure*}[t!]
    \centering
    \begin{subfigure}{0.45\textwidth}
    \centering
    \includegraphics[width=0.9\textwidth] {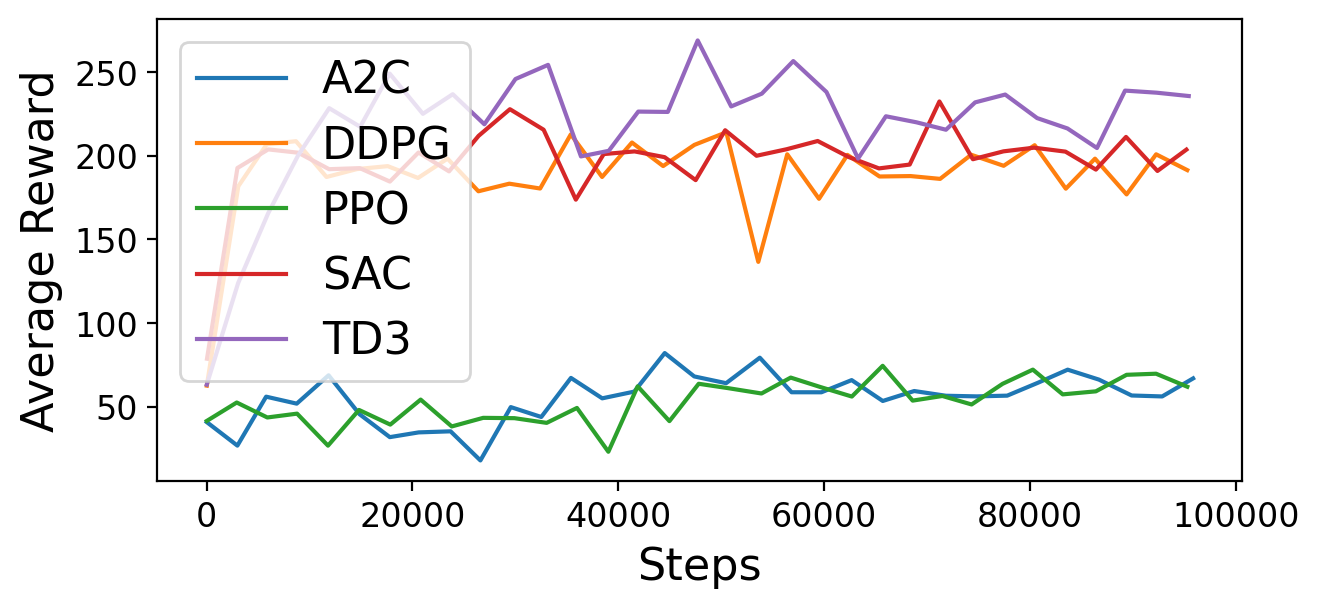}
    \caption{DC Motor $Reach$ task}
    \end{subfigure}
    \begin{subfigure}{0.45\textwidth}
    \centering
    \includegraphics[width=0.9\textwidth] {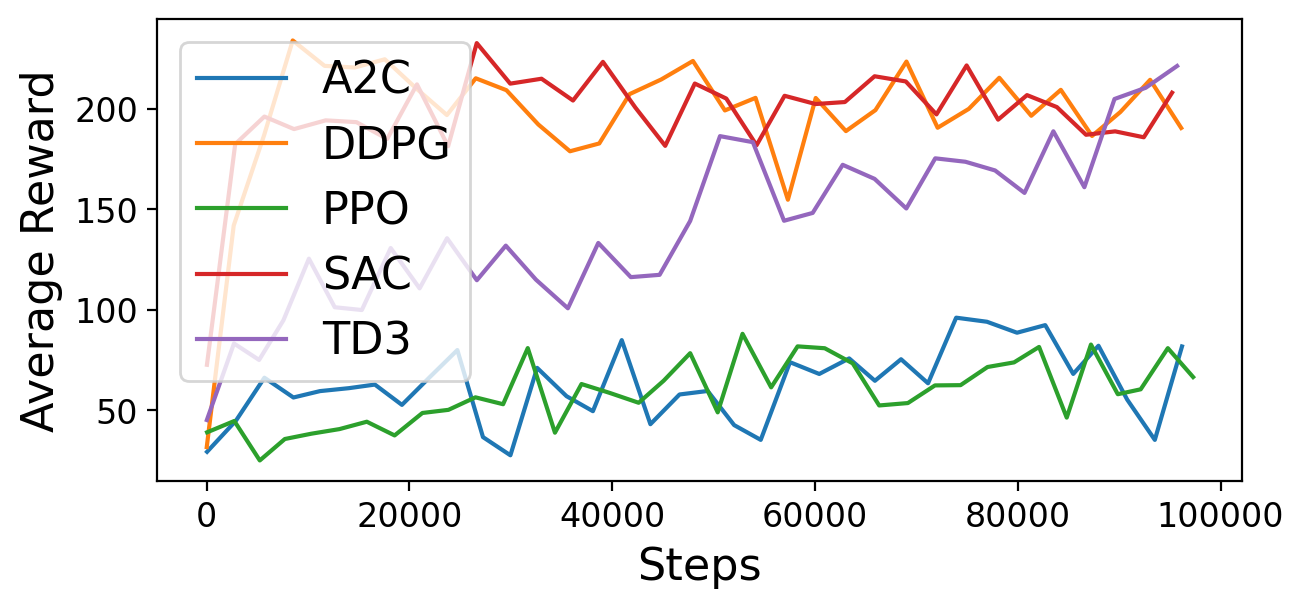}
    \caption{DC Motor $Reach\&Avoid$ task}
    \end{subfigure}
    \begin{subfigure}{0.45\textwidth}
    \centering
    \includegraphics[width=0.9\textwidth] {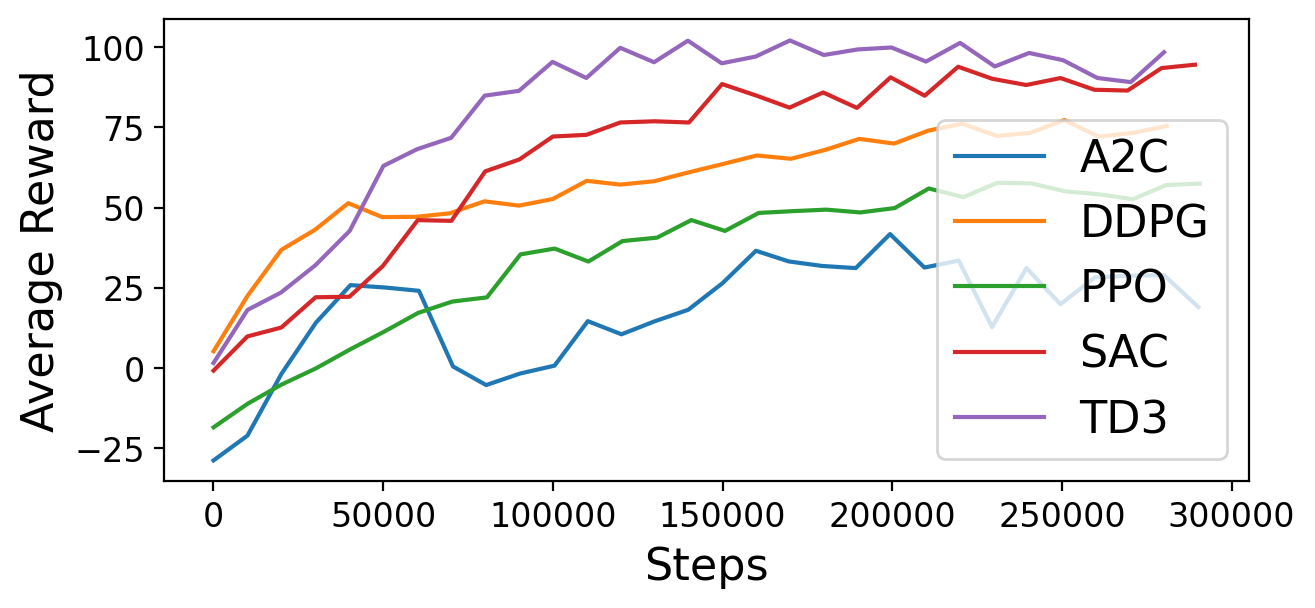}
    \caption{Bicycle $Reach$ task}
    \end{subfigure}
    \begin{subfigure}{0.45\textwidth}
    \centering
    \includegraphics[width=0.9\textwidth] {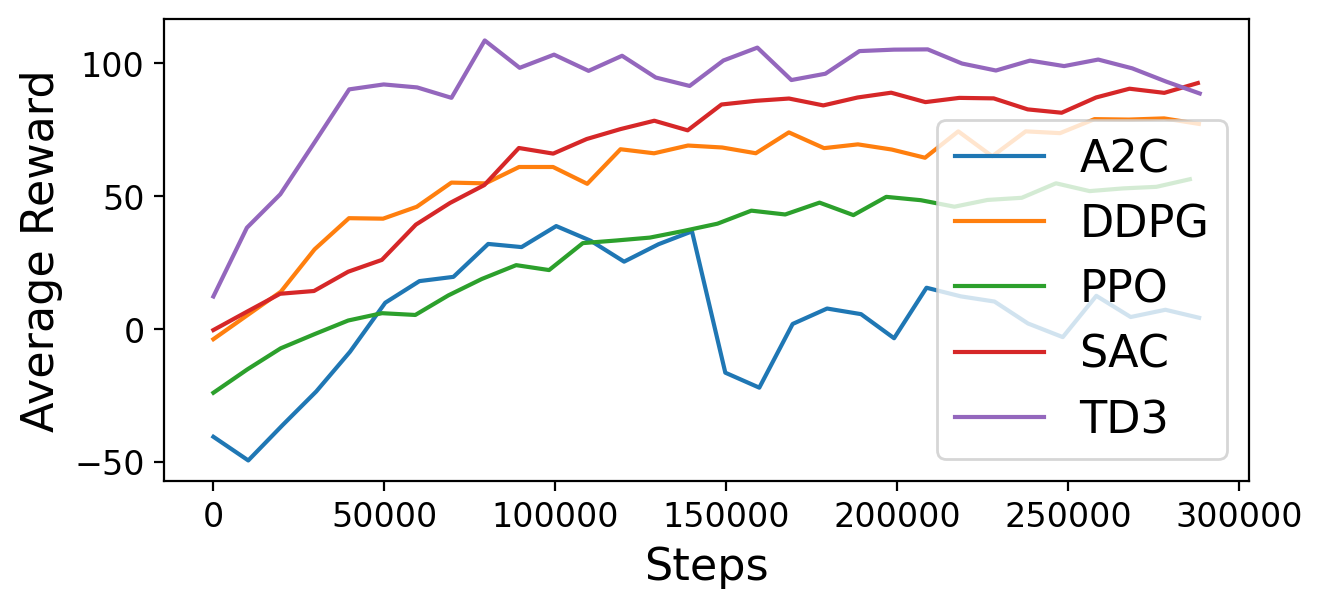}
    \caption{Bicycle $Reach\&Avoid$ task}
    \end{subfigure}
    \begin{subfigure}{0.45\textwidth}
    \centering
    \includegraphics[width=0.9\textwidth] {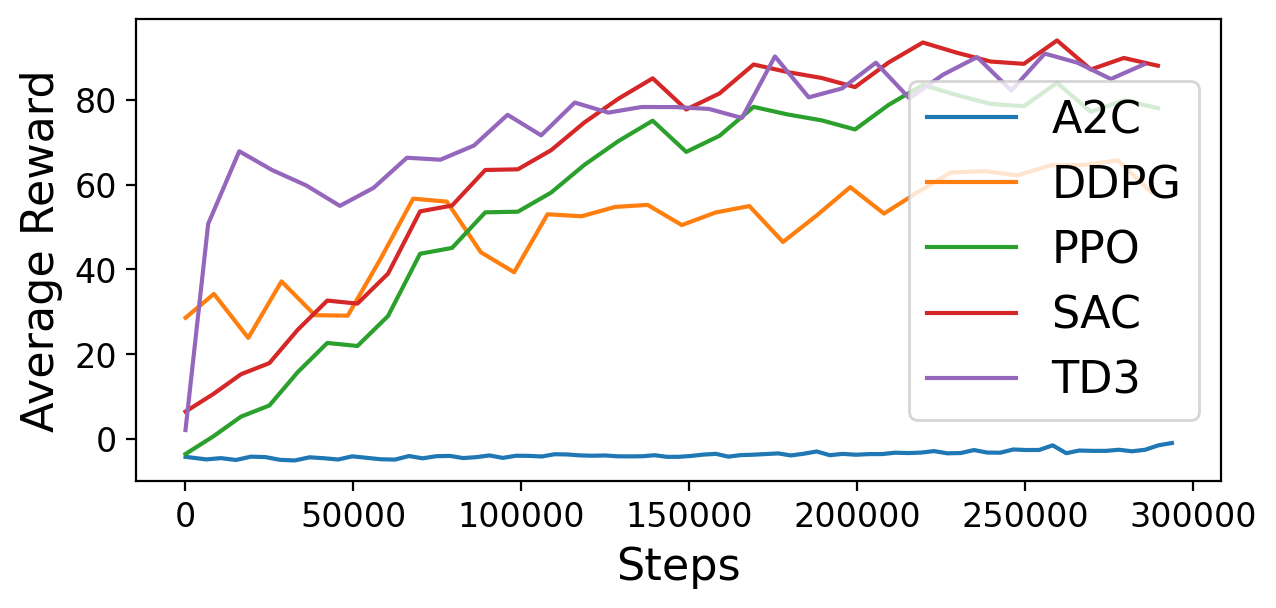}
    \caption{Attitude Control $Reach$ task}
    \end{subfigure}
    \begin{subfigure}{0.45\textwidth}
    \centering
    \includegraphics[width=0.9\textwidth] {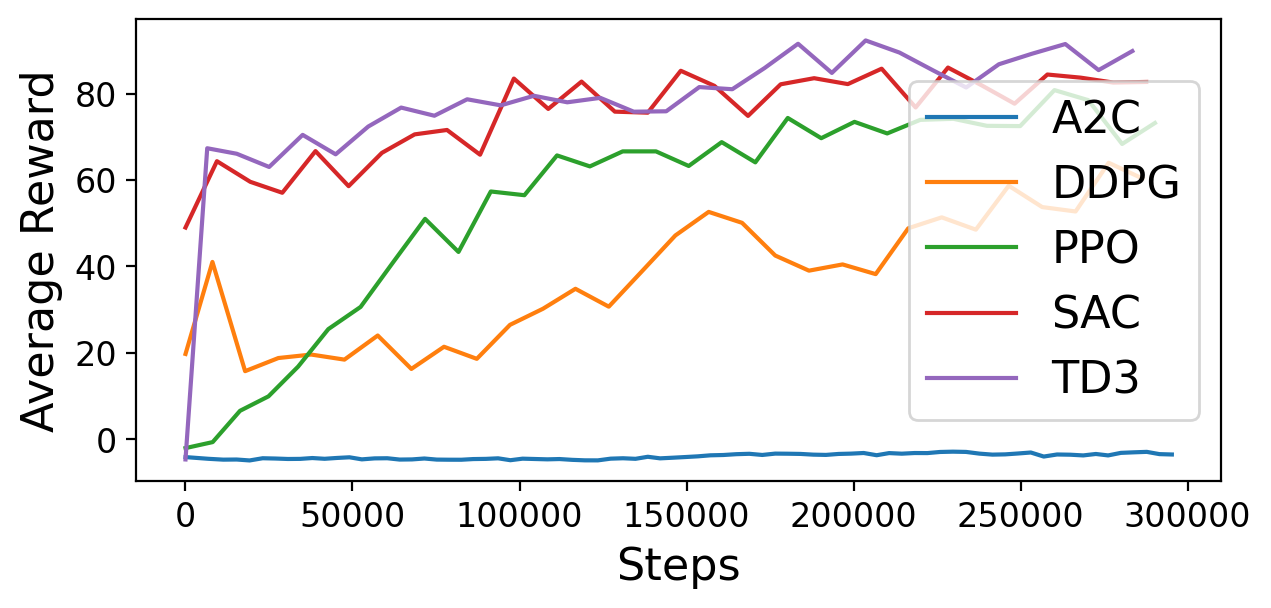}
    \caption{Attitude Control $Reach\&Avoid$ task}
    \end{subfigure}
    \begin{subfigure}{0.45\textwidth}
    \centering
    \includegraphics[width=0.9\textwidth] {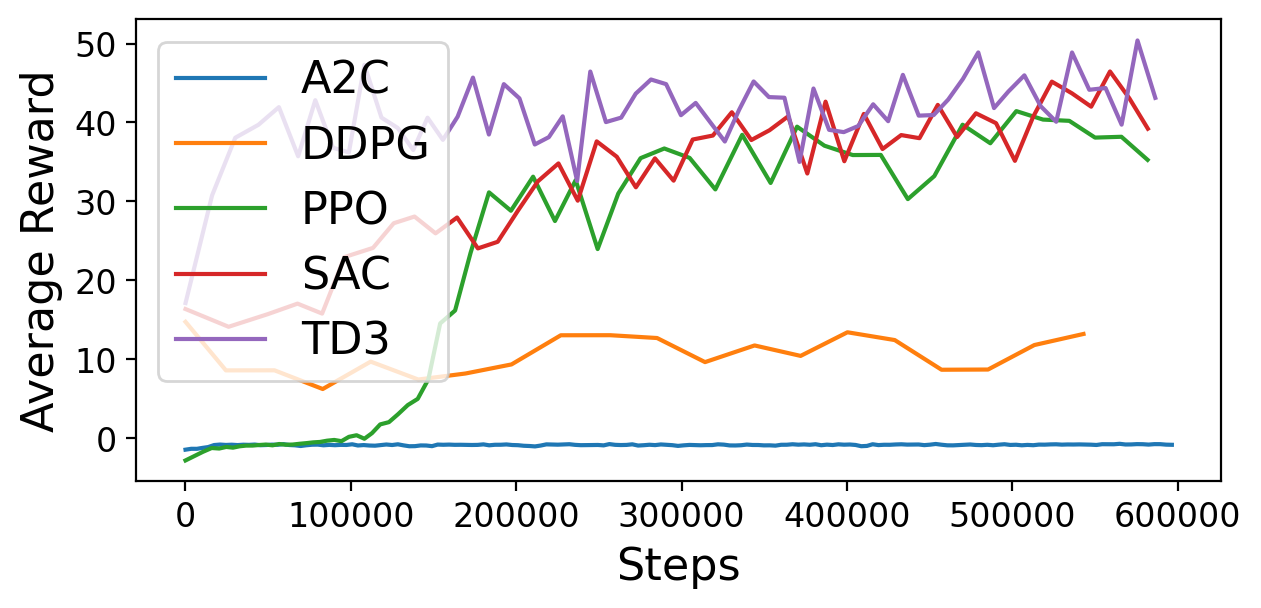}
    \caption{Swimmer $Reach$ task}
    \end{subfigure}
    \begin{subfigure}{0.45\textwidth}
    \centering
    \includegraphics[width=0.9\textwidth] {figs/swimmer_avoid.png}
    \caption{Swimmer $Reach\&Avoid$ task}
    \end{subfigure}
    \caption{Progression of average reward during training on different tasks.}
    \label{fig:training_results}
\end{figure*}

\begin{figure*}[!h]
        \begin{subfigure}[b]{0.25\textwidth}
                \includegraphics[width=\linewidth]{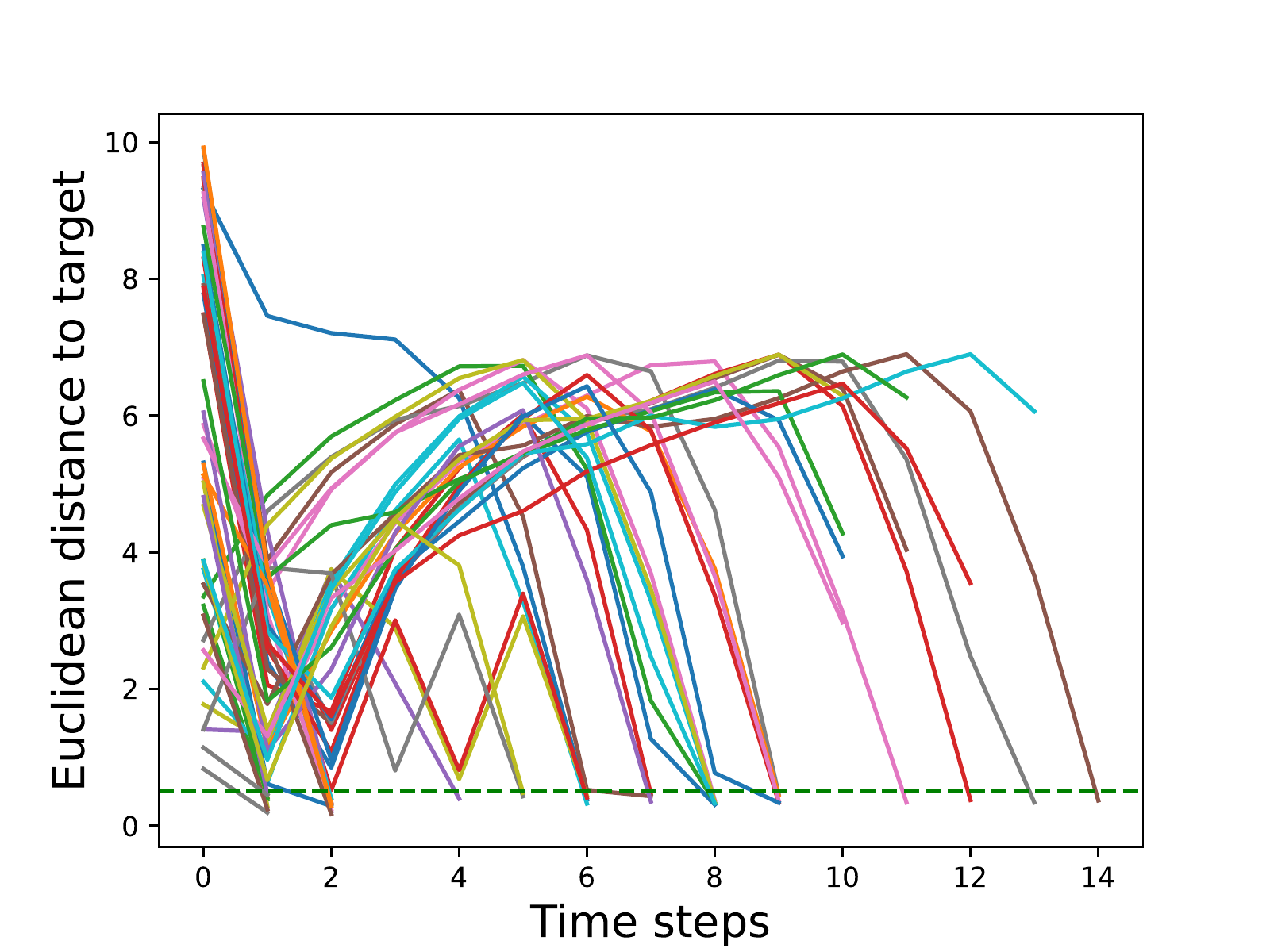}
                \caption{Euclidean Distance}
        \end{subfigure}%
        \begin{subfigure}[b]{0.25\textwidth}
                \includegraphics[width=\linewidth]{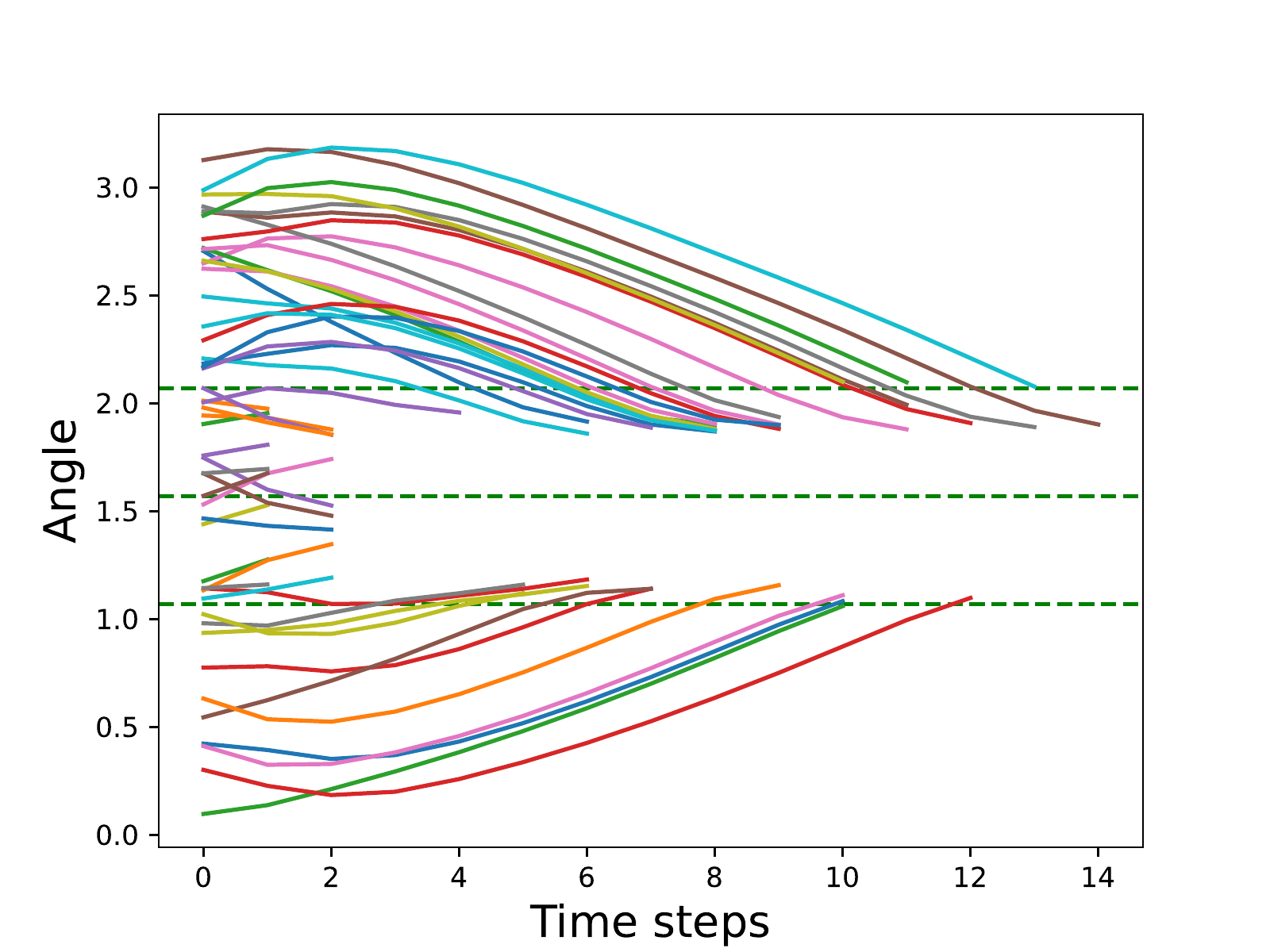}
                \caption{Angle}
        \end{subfigure}%
        \begin{subfigure}[b]{0.25\textwidth}
                \includegraphics[width=\linewidth]{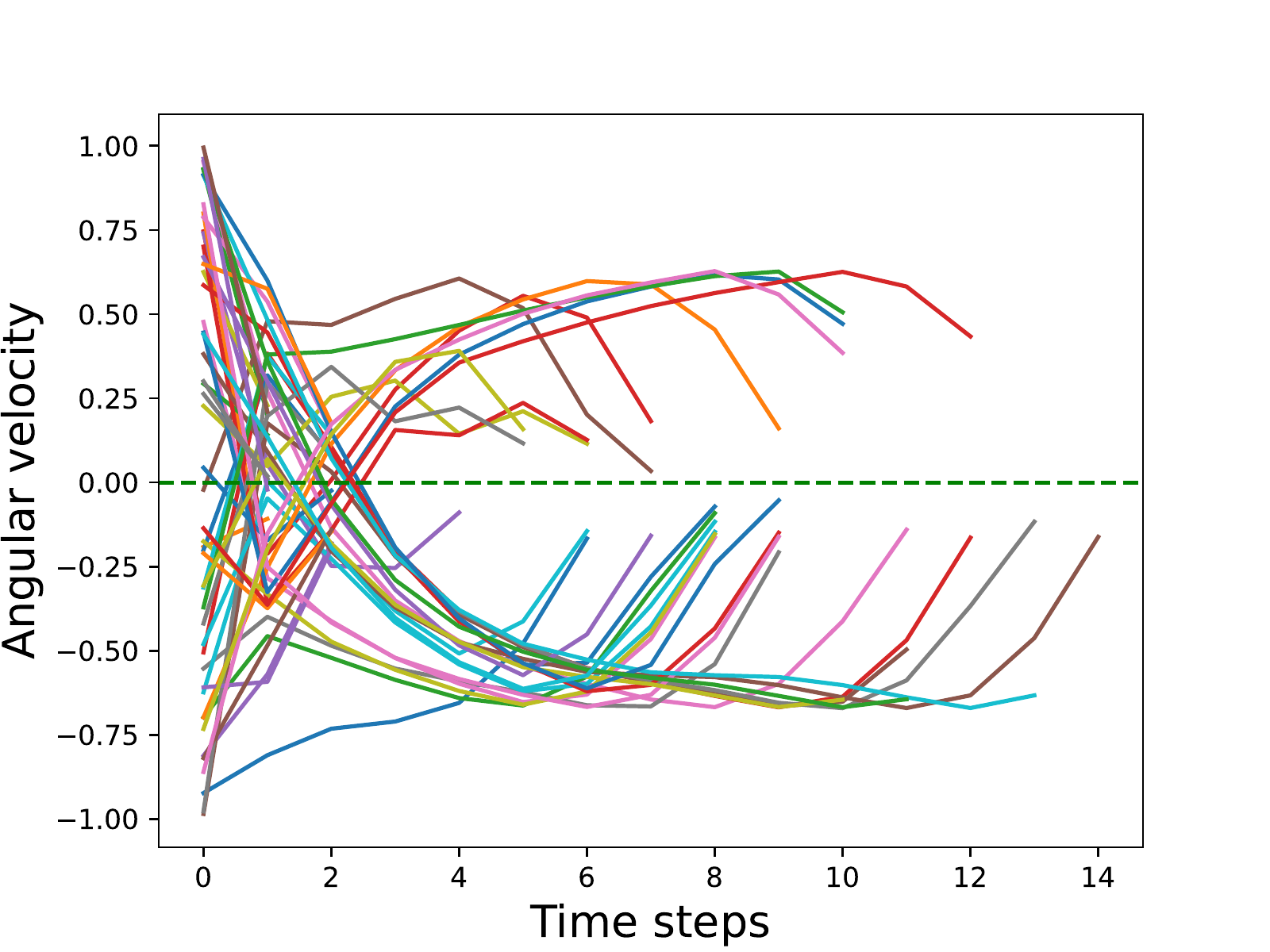}
                \caption{Angular Velocity}
        \end{subfigure}%
        \begin{subfigure}[b]{0.25\textwidth}
                \includegraphics[width=\linewidth]{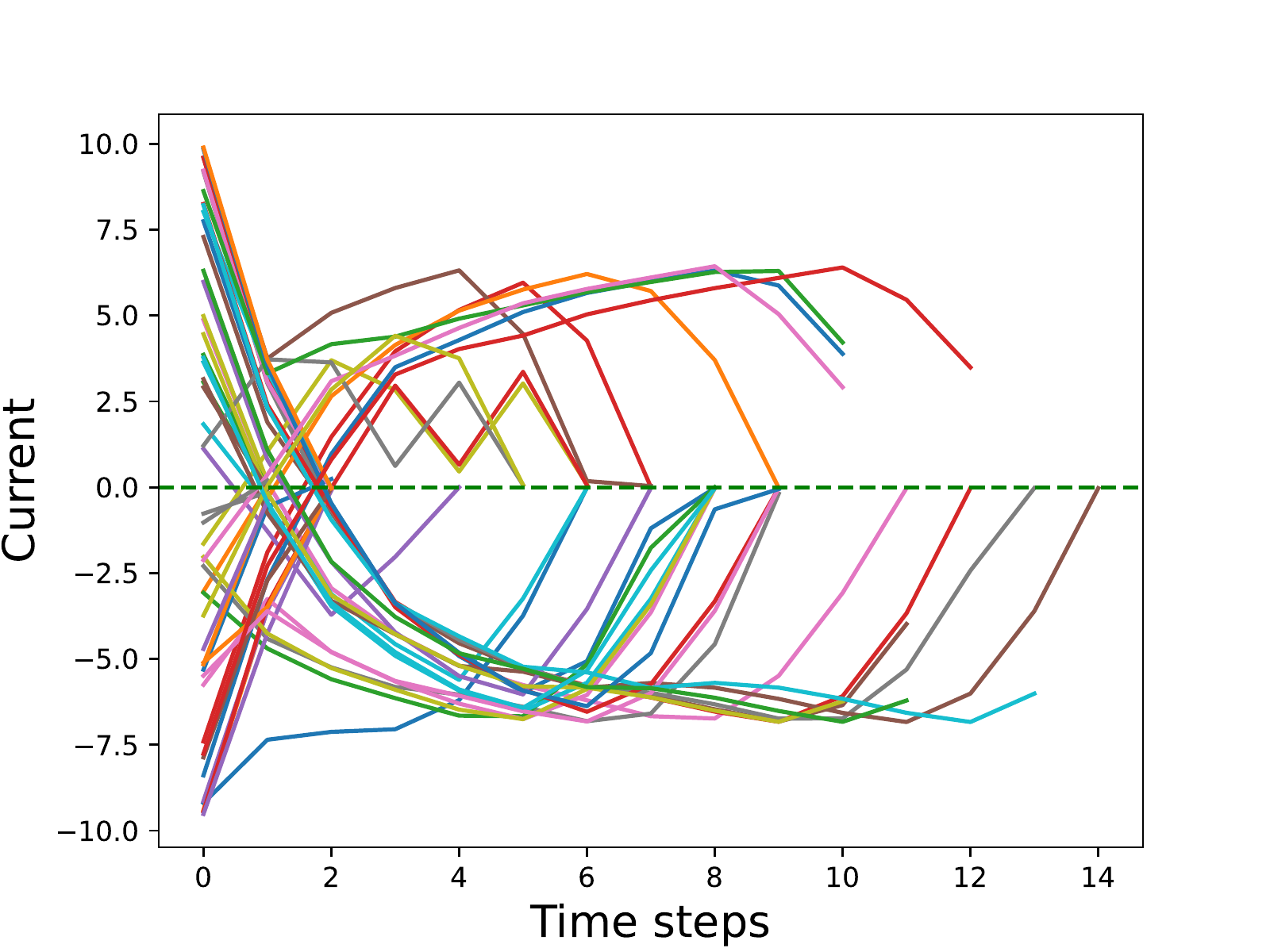}
                \caption{Current}
        \end{subfigure}
        \caption{Example testing trajectories of $Reach$ Task on DC Motor Position Benchmark}\label{fig:example}
\end{figure*}
 The STL specification format for Reach\&Avoid tasks  as follow:
\begin{equation}
\begin{aligned}
    F_{[0,10]}(Euclid(target)\leq r_{t}\ and \\  G_{[0,\infty]}(Euclid(unsafe)>r_{u})
\end{aligned}
\end{equation}
The former part of the formula is $\varphi_T$ for reach task and latter part is $\varphi_O$ for avoid task as defined in section~\ref{subsec:asap_recovery}. We want to achieve $\property_T$ ASAP. $Euclid$ is the euclidean distance on the state space, $r_t$ is the radius of the target set, $r_u$ is the radius of the unsafe set which are specifies in Table \ref{tab:benchmark}.

\section{Detailed Experiment Results}
\label{sec:detailed_experiment_results}
In this section, we show more detailed experiment results for ablation studies.

\subsection{Ablation Study 1}
The series figures showed the training history of different algorithms for different tasks on different benchmarks.

On DC Motor Position benchmark, the training takes 1e6 steps and TD3, SAC and DDPG have similar performance and outperform PPO and A2C.

On Attitude Control benchmark and Bicycle benchmark, the training takes 3e6 steps. The average reward is still increasing after 1e6 as Fig.~\ref{fig:training_results} shows. This echos the intuition we have in appendix A.

On swimmer benchmark, the the training takes 6e6 steps. SAC and TD3 outperform other algorithms.

\subsection{Ablation Study 2}
In this subsection, we discuss alternative ways to achieve ASAP and their performance.
Intuitively, greedy algorithm can solve this problem. We can force the system to move closer and closer to the target set on each step. However, this method is not guaranteed to work since the system dynamics follow the physical law. 

Fig.~\ref{fig:example} shows an example of the change of euclidean distance to the target and system states for a reach task. The goal is to reach $[\pi/2, 0,0,0]$ ASAP. There are 50 trajectories on each figure, the euclidean distance is not monopoly decreasing since the transition of current and angular velocity are required to drive the angle to target, which is determined by the system dynamics. Respectively, the success rate of greedy algorithm for the $Reach$ task on the DC Motor Position benchmark is $67.6\%$. Additionally, it is hard to design how much each step the system is closer to the target set compare to the previous step(step movement constant) since the system dynamics is not known.

Another way to express ASAP is transform the greedy algorithm as a STL-specification which turns to encourage the system to move closer and closer to the target. However, STL does not have the capability to express ASAP and the 
the success rate of greedy algorithm for the $Reach$ task on the DC Motor Position benchmark is $55.3\%$.

For the STL-specification of reach, we use finally operator ($F$) and corresponding semantics. In natural language, the meaning of this specification is "finally reach the target set". It is also possible to change it to the globally operator ($G$) and the trained agents have similar performance.

To be noticed, quantitative semantics of STL is not our contribution. And we have tested if the we use the above alternative ways and ASAP-Phi together on the same task, it can achiever success rate over $95\%$. This result partially shows the necessity of our framework.


\section{Explanation of the Main Proof}

In this appendix section, we further explain the proof of Theorem \ref{thm:asap_phi_correctness} in Section \ref{subsec:analysis} with illustrations.

\begin{figure}[H]
    \centering
    \includegraphics[width=0.5\textwidth]{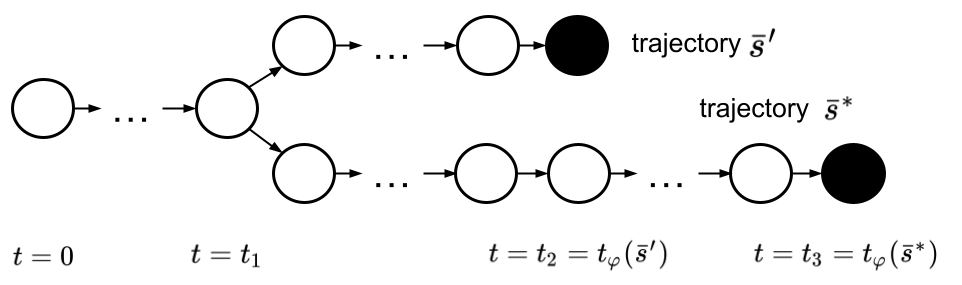}
    \caption{The two existing trajectories under the assumed $\policy^*$. An empty circle represents the trajectory so far does not satisfy $\property$, while a filled circle means satisfy.}
    \label{fig:proof_trajectories_redrawn}
\end{figure}

Overall, the proof aims to show that optimality of a policy of ASAP-Phi implies that it is an ASAP policy. Equivalently, we can show that the existence of an optimal and non-ASAP policy implies a contradiction. Therefore, we assume for the existence of such a policy $\pi^*$, which favors a trajectory $\signal^*$ that satisfies the property later than an alternative trajectory $\signal'$ due to its non-ASAP nature. These two trajectories are illustrated back in Section \ref{subsec:analysis}, and we re-draw it here again for convenience as Figure \ref{fig:proof_trajectories_redrawn}.

\begin{figure}[H]
    \centering
    \includegraphics[width=0.5\textwidth]{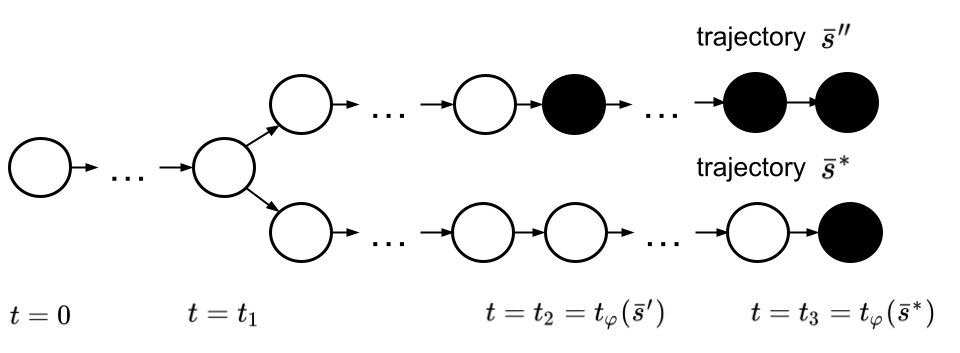}
    \caption{Trajectory $\signal''$ extends from $\signal'$ and keeps property satisfaction from $t_2$ to $t_3$.}
    \label{fig:proof_trajectories_2}
\end{figure}

Next, because of Assumption 1, it is feasible to maintain $\signal'$ satisfactory until time step $t_3$. That is, a trajectory $\signal''$ that extends $\signal'$ until $t_3$, and it keeps property satisfaction all the way from $t_2$ to $t_3$. This is illustrated in Figure \ref{fig:proof_trajectories_2}.

\begin{figure}[H]
    \centering
    \includegraphics[width=0.5\textwidth]{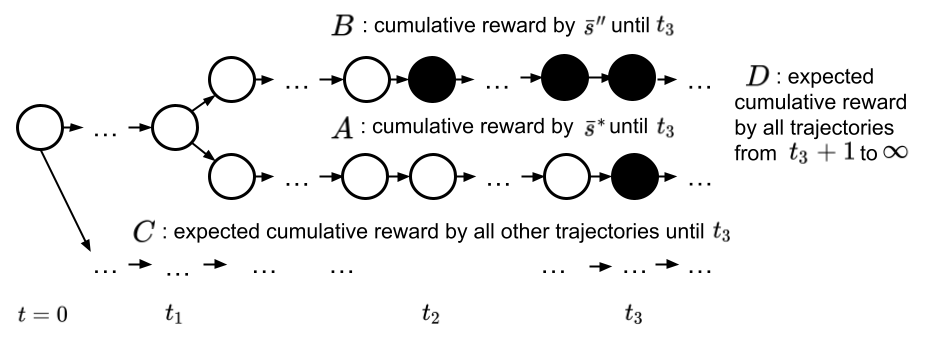}
    \caption{Value at $\state_0$ is split into four parts: $p_{\signal^*}A$, $p_{\signal''}B$, $C$ and $D$.}
    \label{fig:proof_trajectories_3}
\end{figure}

To show that the current policy $\policy^*$ that favors $\signal^*$ more than $\signal'$ (and therefore more than $\signal''$), we need to show that it does not maximize the state value function on all states. Specifically, we pick the common initial state $\state_0$ of both the traces, and  compute the value at this state under policy $\policy^*$. Overall, the value can be split into four portions, as illustrated in Figure \ref{fig:proof_trajectories_3}.

The split is deduced by Equations \eqref{eq:split_1} and \eqref{eq:split_2} in the proof. Overall, denote the probability (under some policy) of taking trajectory $\signal^*$ and $\signal''$ as $p_{\signal^*}$ and $p_{\signal''}$, respectively. Regardless of the policy, the four portions are

\begin{enumerate}
    \item The expected cumulative reward from taking trajectory $\signal^*$ from beginning to $t_3$, i.e., $p_{\signal^*}A$,
    \item The expected cumulative reward from taking trajectory $\signal''$ from beginning to $t_3$, i.e., $p_{\signal''}B$,
    \item The expected cumulative reward from taking any other trajectory aside from these two, from beginning to $t_3$, i.e., $C$, and
    \item The expected cumulative reward from taking any trajectory from $t_3+1$ to $\infty$, i.e., $D$.
\end{enumerate}

Therefore, for any policy $\policy$, we have the value at state $\state_0$ as
\begin{equation}
    \criticval^*_\policy(\state_0) = p_{\signal^*}A + p_{\signal''}B + C_\policy + D_\policy.
\end{equation}
Notice that $A$ and $B$ are deterministic terms and are independent of any policy distribution, but $C$ and $D$ are expected values, so they are parameterized by the distribution $\policy$.

Under our assumed optimal policy $\policy^*$, the policy favors $\signal^*$ more than $\signal''$, i.e., $p_1 = p_{\signal^*} > p_{\signal''} = p_2$ under $\policy^*$, we have
\begin{equation}
    \criticval^*_{\policy^*}(\state_0) = p_1A + p_2B + C_{\policy^*} + D_{\policy^*}.
\end{equation}

After finding this split of value, we seek an alternative policy $\policy'$, which has a higher value $\criticval^*_{\policy'}(\state_0) > \criticval^*_{\policy^*}(\state_0)$, and therefore $\policy^*$ does not maximize the value at state $\state_0$ and is not optimal, so that a contradiction occurs.

As stated in the main proof, we construct the alternative policy $\policy'$ by swapping the probabilities of taking trajectories $\signal''$ and $\signal^*$, while the probabilities on everything else remain the same. This give us $p_{\signal''} = p_1$ and $p_{\signal^*} = p_2$, i.e.,
\begin{equation}
    \criticval^*_{\policy'}(\state_0) = p_2A + p_1B + C_{\policy'} + D_{\policy'}.
\end{equation}

From Equation \eqref{eq:alternative_value} to the end of the proof, we compare the two values term-by-term. This procedure indeed shows that $\criticval^*_{\policy'}(\state_0) > \criticval^*_{\policy^*}(\state_0)$ and reaches the contradiction.

\end{appendices}